\documentclass[pdflatex,sn-mathphys]{sn-jnl}
\jyear{2022}

\newtheorem*{definition}{Definition} 
\newtheorem*{assumption}{Assumption} 
\newtheorem*{lemma}{Lemma} 
\newtheorem*{theorem}{Theorem}

\newcommand{\tname}{\texttt{CCTS}\xspace}
\newcommand{\mname}{\texttt{RU}\xspace}

\usepackage{lineno} 

\begin{document}


\title[CCTS]{Continuous Diagnosis and Prognosis by Controlling the Update Process of Deep Neural Networks}

\author[1,2]{\fnm{Chenxi} \sur{Sun}}\email{sun\_chenxi@pku.edu.cn} 

\author*[1,2]{\fnm{Hongyan} \sur{Li}}\email{leehy@pku.edu.cn}

\author[1,2]{\fnm{Moxian} \sur{Song}}\email{songmoxian@pku.edu.cn}

\author[1,2]{\fnm{Derun} \sur{Cai}}\email{cdr@stu.pku.edu.cn}

\author[1,2]{\fnm{Baofeng} \sur{Zhang}}\email{boffinzhang@stu.pku.edu.cn}

\author*[3,4]{\fnm{Shenda} \sur{Hong}}\email{hongshenda@pku.edu.cn}

\affil[1]{\orgdiv{Key Laboratory of Machine Perception (Ministry of Education)}, \orgname{Peking University}, \orgaddress{ \city{Beijing}, \country{China}}}

\affil[2]{\orgdiv{School of Intelligence Science and Technology}, \orgname{Peking University}, \orgaddress{ \city{Beijing}, \country{China}}}

\affil[3]{\orgdiv{National Institute of Health Data Science}, \orgname{Peking University}, \orgaddress{ \city{Beijing}, \country{China}}}

\affil[4]{\orgdiv{Institute of Medical Technology}, \orgname{Health Science Center of Peking University}, \orgaddress{ \city{Beijing}, \country{China}}}

\abstract{
Continuous diagnosis and prognosis are essential for intensive care patients. It can provide more opportunities for timely treatment and rational resource allocation, especially for sepsis, a main cause of death in ICU, and COVID-19, a new worldwide epidemic. Although deep learning methods have shown their great superiority in many medical tasks, they tend to catastrophically forget, over fit, and get results too late when performing diagnosis and prognosis in the continuous mode. In this work, we summarized the three requirements of this task, proposed a new concept, continuous classification of time series (\tname), and designed a novel model training method, restricted update strategy of neural networks (\mname). In the context of continuous prognosis, our method outperformed all baselines and achieved the average accuracy of 90\%, 97\%, and 85\% on sepsis prognosis, COVID-19 mortality prediction, and eight diseases classification. Superiorly, our method can also endow deep learning with interpretability, having the potential to explore disease mechanisms and provide a new horizon for medical research. We have achieved disease staging for sepsis and COVID-19, discovering four stages and three stages with their typical biomarkers respectively. Further, our method is a data-agnostic and model-agnostic plug-in, it can be used to continuously prognose other diseases with staging and even implement \tname in other fields.
}


\maketitle

\section*{Introduction} \label{sec:introduction}

Continuous diagnosis and prognosis are of great significance for timely, personalized treatment and rational allocation of medical resources. Especially in the Intensive Care Unit (ICU), the status perception and disease diagnosis are needed at any time as the real-time diagnosis provides more opportunities for doctors to rescue lives \cite{2014The}. For example, sepsis is a life-threatening condition, causing more than half of ICU deaths \cite{sepsis-3}. Early detection and antibiotic treatment are critical for improving sepsis outcomes \cite{2017Time}; Corona Virus Disease 2019 (COVID-19) outbreaks have caused health concerns worldwide \cite{covid-19who}. In the case of a sudden outbreak of the new epidemic, the continuous prognosis can help for personalized treatment and rational allocation of scarce resources \cite{COVID-19}.

\begin{figure*}[!t]
\centering
\includegraphics[width=\linewidth]{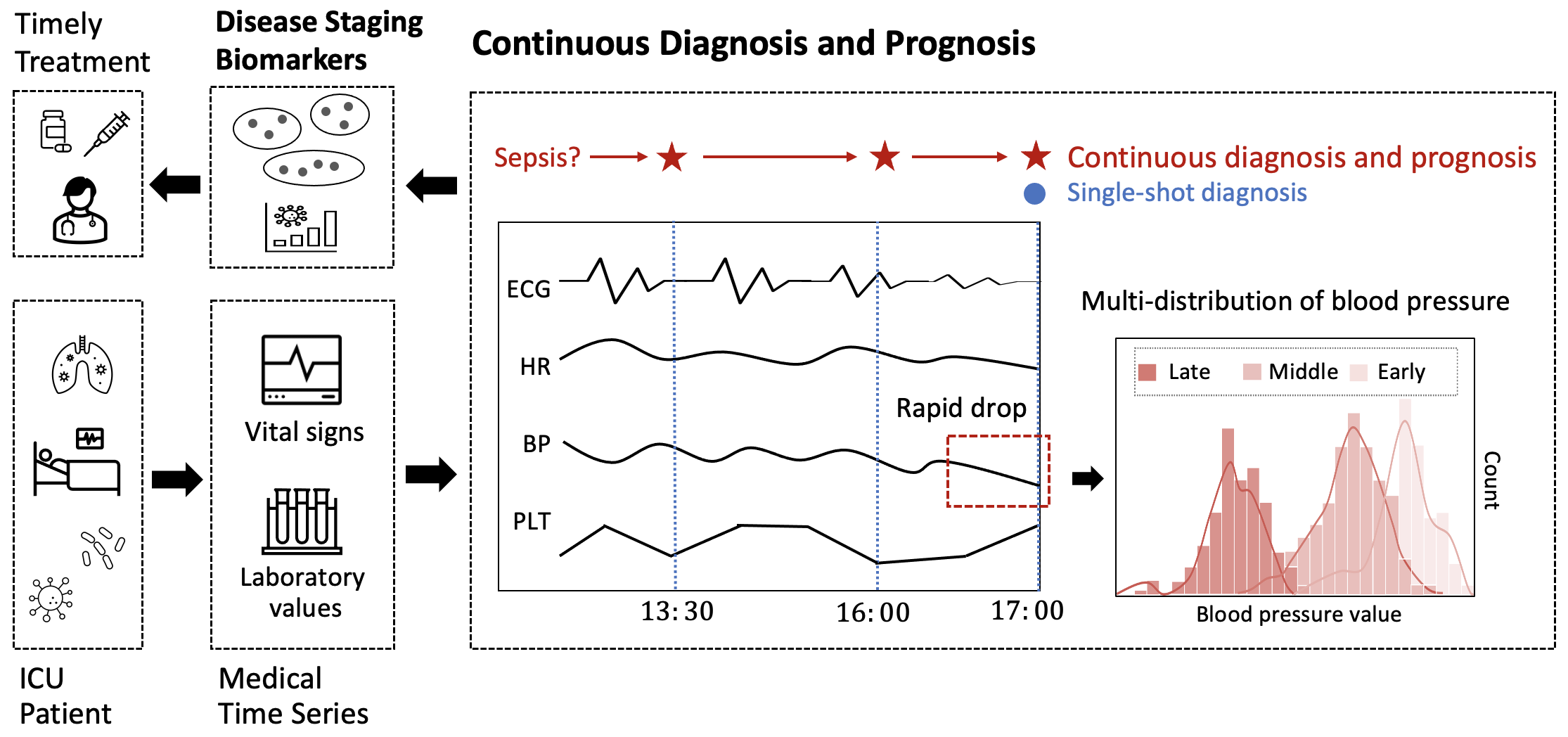}
\caption{Continuous Diagnosis and Prognosis with Disease Staging}
\label{fig:introduction}
\end{figure*}

Different from the single-shot diagnosis, which is often made for the outpatient, the task of continuous diagnosis and prognosis emphasizes the multiple early diagnoses or prognosis for the inpatient at different stages over time. For example, in Figure \ref{fig:introduction}, an ICU patient is monitored for vital signs in real-time. Assuming that he will be in sepsis shock at 17:00, the common diagnostic system will give a warning when he is suffering or about to suffer from sepsis at about 17:00 (the single-shot diagnosis, blue dot). This is likely to miss the emergency treatment time for the acute disease, where each hour of delay has been associated with roughly a 4-8\% increase in sepsis mortality \cite{2017Time}. Thus, we require the continuous prognosis for sepsis (red stars), where we can predict the patient outcome 4 hours early, 1 hour early, etc. at 13:00, 16:00, etc. In order to meet the practical need, we summarized four requirements in the task of continuous diagnosis and prognosis.

\textbf{Requirement 1: the ability to identify symptoms in different time stages before the disease onset.} The single-shot diagnosis only needs to learn the clinical manifestation, which is easy under the guidance of the gold standard \cite{Mariam2008Rapid}. But the continuous prognosis needs to learn the underlying symptoms of the disease, which are usually not obvious in the clinic and cannot be judged by clinicians. And the symptoms are not only from a certain stage but from multiple stages before the onset, leading to diversity and hybridity.

\textbf{Requirement 2: potential for earlier diagnosis with guaranteed accuracy.} Earlier diagnosis is necessary for many severe illnesses. E.g., each hour of delayed treatment could cause a 4-8\% increase in sepsis mortality \cite{2017Time}. But basic questions about the limits of early detection remain unanswered. If one wants to pursue higher diagnostic accuracy, it would tend to predict late for clearer features. E.g., the rapid drop of blood pressure (a major symptom of sepsis shock, the red dashed box in Figure \ref{fig:introduction}) always occurs just before the shock \cite{singer2016third}. Thus, we expect the continuous mode to achieve earlier and more accurate results than the single-shot mode.

\textbf{Requirement 3: merits of explainability and disease staging.} The 22nd article of the European Union’s General Data Protection Regulation stipulates that a subject of algorithmic decisions has a right to meaningful explanation regarding said decisions \cite{2016EU}. As clinicians always justify a result using medical-domain knowledge familiar to them \cite{DBLP:conf/mlhc/TonekaboniJMG19}, the explainable methods will be more popular in practice. 

Meanwhile, the continuous prognosis is accompanied by the disease progression. Disease staging is important to understand disease mechanisms and implement targeted treatment. A clinically useful staging system stratifies patients by their baseline risk of an adverse outcome and their potential to respond to therapy. The best developed and most explicit approach has evolved in oncology \cite{cancerstage}, but it is not clear for critical illnesses. E.g., the stratification of sepsis, severe sepsis, and septic shock is questioned in the latest sepsis definition \cite{sepsis-3} and there are no criteria for temporal septic stages \cite{20032001}.

\textbf{Requirement 4: function of offline and sustainable use. }
In many scenarios, especially in ICU, we need to directly use the mature system without constant adjustment. A well-informed system can reduce the risk of misjudgment \cite{DBLP:conf/icml/SiZZB20}. Further, in subsequent applications, when obtaining a batch of new data, such as new patients and new clinical observations, we hope to continue to use the current system instead of designing a new one. Because the new system cannot handle the old data well, the data accord with the old knowledge may still be generated. 

Nowadays, many studies have shown that Deep Learning (DL) methods \cite{2015Deep} are superior to medical gold standards and experienced doctors in some medical tasks such as medical image recognition \cite{0Deep} and arrhythmia detection \cite{hannun2019cardiologist}. Surprisingly, in these studies, sequential medical records, such as vital signs, multiple blood samples, and serial medical imaging, provided more possibilities for DL models to implement diagnosis and prognosis. We uniformly name such sequential medical records as medical time series data. However, most DL-based models often give the single-shot diagnosis after learning the full-length medical time series, but can not prognose continuously. Although some sub-disciplines also study the mode of continuous learning (Appendix \ref{sec:related work}). But they cannot satisfy the above requirements at the same time.

The requirements ask the DL model to learn multi-distributed medical data with interpretability. The labels (mortality, morbidity, etc.) of the real-world medical time series are usually determined at the final time. If the model simply learns the full-length time series, it can only give the single-shot result at the onset time. For continuous diagnosis and prognosis, the model needs to learn time series from different advanced stages: When the data changes, the model performance needs to maintain. But most medical time series have evolved distribution. In Figure \ref{fig:introduction}, the blood pressure varies among early, middle, and late stages, bringing a triple-distribution. DL models are lack of ability to learn all distributions simultaneously due to the premise of independently and identically distribution \cite{DBLP:conf/aaai/ShimMJSKJ21}. Learning new knowledge could inevitably lead to the forgetting of old ones \cite{DBLP:journals/nn/ParisiKPKW19}, and learning one distribution frequently may fall into the local solution with overfitting \cite{DBLP:conf/iclr/SahaG021}. Meanwhile, unfortunately, interpretability is an elusive concept. The field of artificial intelligence holds no consensus regarding its definition and common opinion holds that DL models are uninterpretable black-boxes \cite{2016Can}.

\begin{figure*}[!t]
\centering
\includegraphics[width=\linewidth]{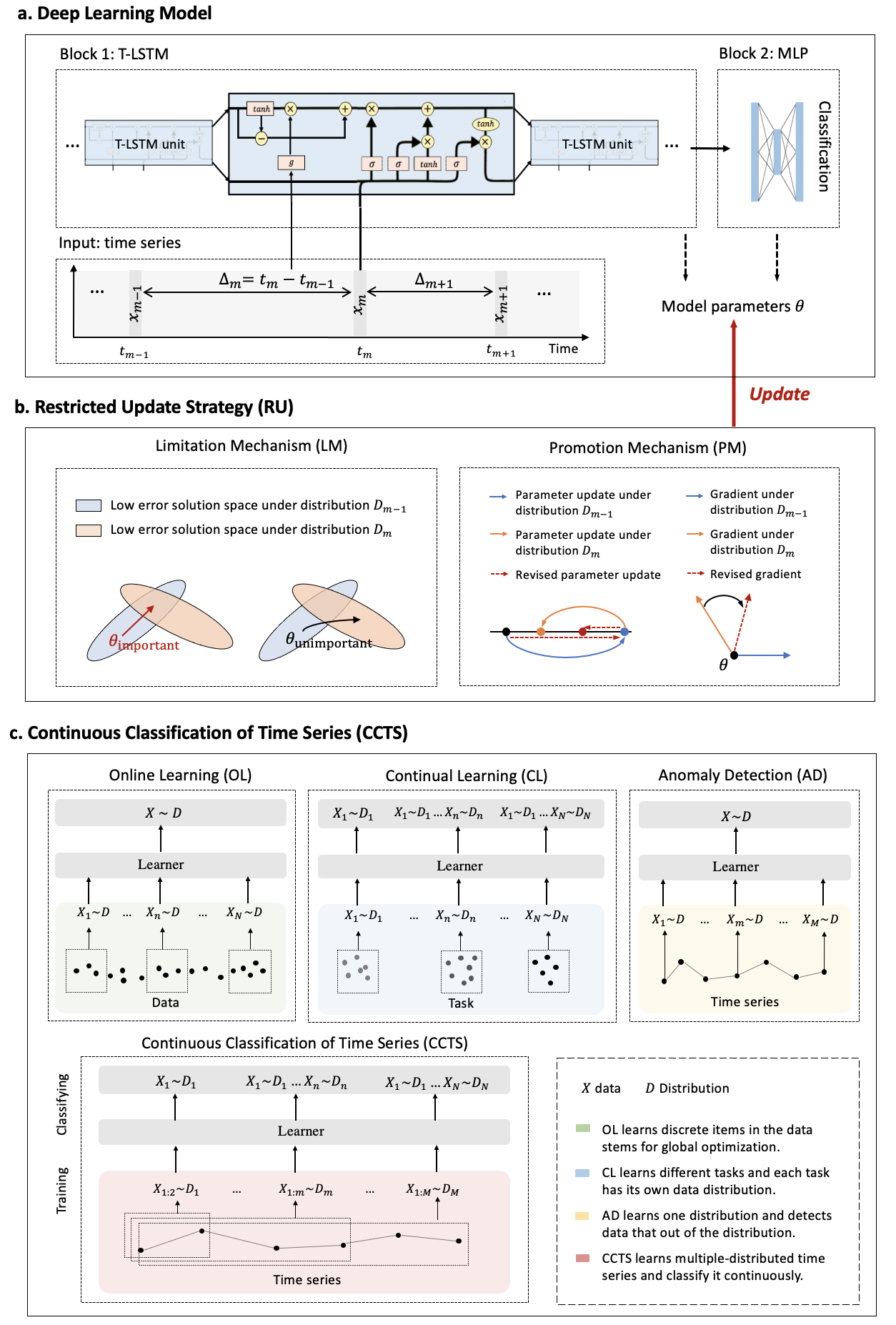}
\caption{Restricted Update Strategy of Neural Networks (\mname) for Continuous Classification of Time Series (\tname)}
\label{fig:method}
\end{figure*}

To this end, we established a novel training method for DL models, Restricted Update strategy of neural network parameters (\mname). \mname can satisfy the above requirements: For Requirement 1, it has the Limitation Mechanism (LM) to avoid catastrophic forgetting and overfitting; For Requirement 2, it has the Promotion Mechanism (PM) to consolidate the knowledge of early distribution; For Requirement 3, we defined the importance coefficient of model parameters to reveal the development of the model and achieve disease staging with typical biomarkers; For Requirement 4, our method trains the DL model by real-world datasets with separate training and test sets and having the test of continual use. Experimental results show that our method is more accurate than all baselines, achieving the accuracy of 90\%, 97\%, and 85\% on sepsis prognosis, COVID-19 mortality prediction, and eight diseases diagnoses. 

The major advantages of our study are fourfold: (1) For continuous diagnosis and prognosis of time-sensitive illness, we design a \mname strategy for the DL model, which outperforms baselines. (2) \mname has a certain ability to interpret the update of DL models and the change of medical time series through input indicators and parameter visualization. These side effects make our method attractive in medical applications where model interpretation and marker discovery are required. (3) We extend our method to connect the distribution change of vital signs with the parameter change of the DL model and we find typical disease biomarkers and stages of sepsis and COVID-19. (4) \mname is a data-agnostic, model-agnostic, and easy-to-use plug-in. It can be used to train various types of DL models. Note that such a continuous prediction mode is needed in most time-sensitive applications, not just in medical tasks. We define these tasks with a new concept, Continues Classification of Time Series (\tname).

\section*{Results}\label{sec2}

\subsubsection*{A restricted update strategy to train deep learning models for continuous diagnosis and prognosis.} 

We sought to develop a training strategy that could help deep learning models to classify time series continuously, especially to achieve continuous medical diagnosis and prognosis. To this end, we focused on continuous sepsis prognosis, continuous COVID-19 mortality prediction, and continuous eight diseases classification based on medical time series, including vital signs from various monitors, and continuous blood sample records during hospitalization. All used data is available: CinC-19 dataset \cite{DBLP:conf/cinc/ReynaJSJSWSNC19} has 30,336 ICU patient records with 2,359 diagnosed sepsis from three separate hospital systems; COVID-19 dataset \cite{COVID-19} has 6,877 blood sample records of 485 COVID-19 patients from Tongji Hospital, Wuhan, China; MIMIC-III dataset \cite{johnson2016mimic} has 19,993 admission records from 7,537 patients. We focus on 8 diseases (Appendix \ref{sec:experiments}). 

A time series dataset $\mathcal{T}=\{X^{n}\}_{n=1}^N$ has $N$ samples. Each sample $X^{n}_{1:M}=\{x^{n}_{m}\}_{m=1}^M$ has $M$ observations with value $x^{n}_{m}$ and time $t^{n}_{m}$. DL models have achieved great success in modeling medical sequential data \cite{LeCun2015, Shenda2020}, especially Recurrent Neural Network (RNN). However, the real-world time series is usually long and irregularly sampled \cite{DBLP:journals/corr/abs-2010-12493}. For example, critically ill patients are often hospitalized for several months, and records often have hundreds of observations. Due to the change in patient's health status, the relevant measurement requirements are also changing, which may be several hours or days apart \cite{ijcai2021-414}. Thus, to model long-term dependency and eliminate the impact of uneven time intervals, we implemented Time-aware Long Short-Term Memory (T-LSTM) \cite{Baytas:2017:PSV:3097983.3097997}. As shown in Figure \ref{fig:method}a, our DL architecture has two blocks: Block 1 uses T-LSTM to model the input data and represent their hidden features with the consideration of time decay $\Delta_{m}=t_{m}-t_{m-1}$; Block 2 uses Multilayer Perceptron (MLP) to map the features to the class of input data.

After using the full-length time series $X_{1:M}$ to train the proposed deep learning model, it can achieve the average accuracy of 92\%, 97\%, and 88\% on single-shot diagnosis for sepsis, COVID-19, and eight diseases. However, when applying it to continuous diagnosis and prognosis, the accuracy drops by more than 15\%. Thus, we used the dataset of subsequences of time series in different stages $\mathcal{T}^{*}=\{X^{n}_{1:m}\}_{n,m=1}^{N,M}$ to train the model. As the time series changes dynamically, $X_{1:m-1}$ and $X_{1:m}$ will form different data distributions $D_{m-1}$ and $D_{m}$. For learning such multi-distribution, we design a restricted update strategy \mname to train our model. As shown in Figure \ref{fig:method}b, \mname has two mechanisms: Limitation Mechanism (LM) and Promotion Mechanism (PM).

LM helps DL models to learn multi-distributed data, alleviating problems of catastrophic forgetting and overfitting. Due to the observation of many parameter configurations resulting in the same performance \cite{DBLP:journals/nn/ParisiKPKW19}, we could add a regular term to the loss to restrict the updating of model parameters. To this end, when learning a new distribution, LM constrains important parameters $\theta_\text{important}$ for the old distribution to stay close to their old values but changes unimportant parameters $\theta_{\text{unimportant}}$ more. As shown in Figure \ref{fig:method}b, when learning distribution $D_{m}$, $\theta_\text{important}$ is limited to the low error space of distribution $D_{m-1}$ while $\theta_\text{unimportant}$ can be updated to other spaces. The importance of parameter $\theta$ is matured by the importance coefficients $\alpha(\theta)$.

PM helps DL models to classify time series earlier in time-sensitive applications. LM regards early distributions and the new distributions as solutions to the same continuous optimization problem \cite{DBLP:journals/corr/abs-1909-05207} with regret minimization. It is projection-free and estimates a stochastic recursive estimator to alleviates the complexity and training instability. As shown in Figure \ref{fig:method}b, when learning distribution $D_{m}$, PM changes the current gradient from an obtuse angle to an acute angle with the gradient on previous distribution $D_{m-1}$. Because when the new gradient and the old gradients are at an acute angle, the model performance on the old distribution will improve, at least not decrease \cite{DBLP:conf/nips/Lopez-PazR17}. The promotion of learning old distributions has the potential for early classification.

\begin{figure*}[!t]
\centering
\includegraphics[width=\linewidth]{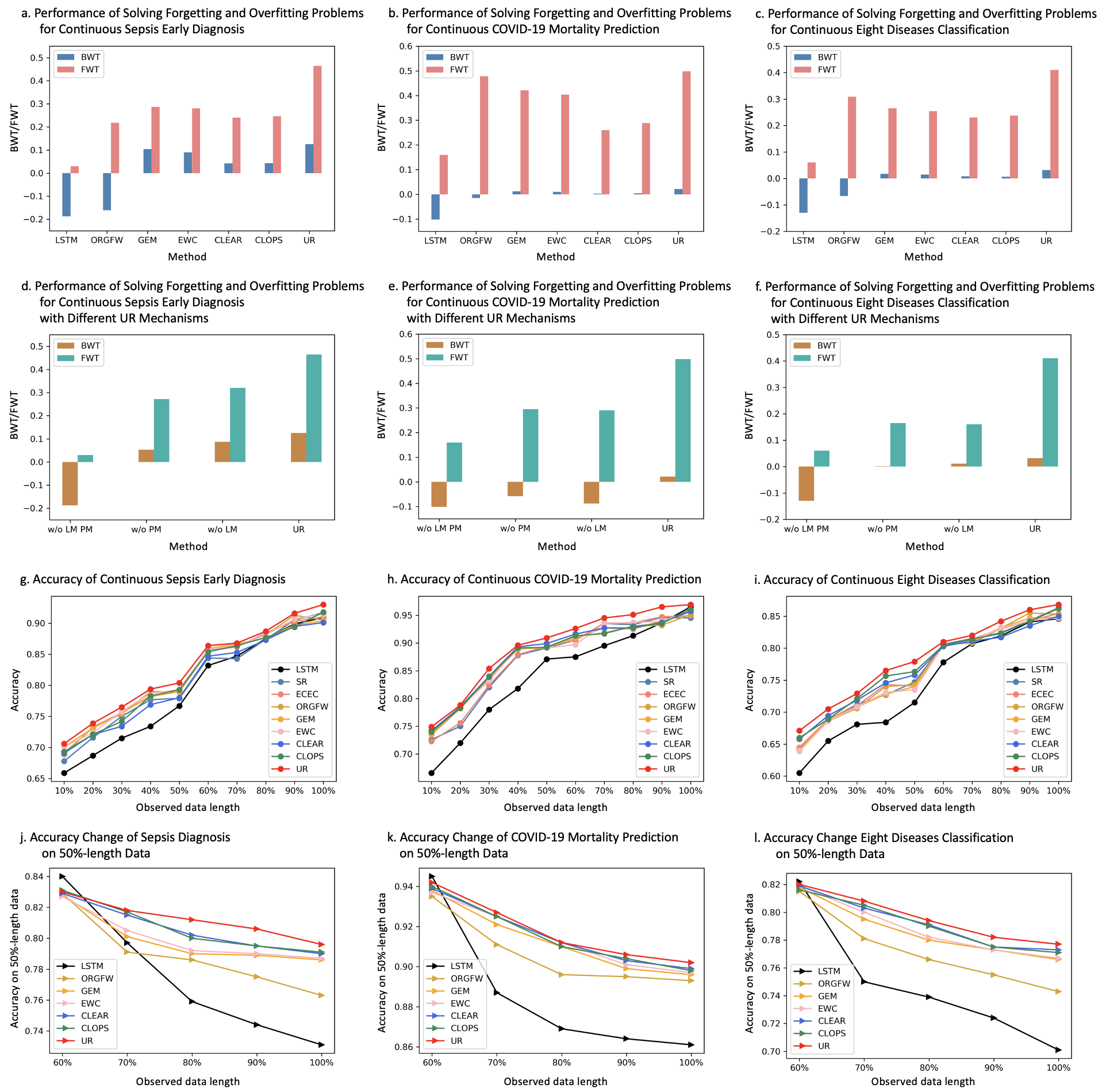}
\caption{Accuracy Results of Continuous Diagnosis and Prognosis}
\label{fig:result1}
\end{figure*}

\subsubsection*{Finding 1: The continuous mode has more potential in medical diagnosis and prognosis than the single-shot mode.} 

As shown in Figure \ref{fig:result1}g-l, our method can classify more accurately at every time. It is significantly better than baselines. In Bonferroni-Dunn test, $k=9, n=3, m=5, q_{0.05}=2.724$ are the number of methods, datasets, cross-validation fold, critical value, then $N=n\times m=15, CD=q_{0.05}\sqrt{\frac{k(k+1)}{6N}}=2.724$, average rank of baselines $\overline{r}=5.5>CD$. Thus, the accuracy is significantly improved. The average accuracy is about 2\% higher, especially in the early time, being 5\% higher for 10\%-length data. 

\tname is important for time-sensitive applications, especially for acute and critical illnesses. Take sepsis diagnosis as an example, compared with the best baseline, our method improves the accuracy by 1.4\% on average, 2.2\% in the early 50\% time stage when the key features are unobvious. Each hour of delayed treatment increases sepsis mortality by 4-8\% \cite{seymour2017time}. With the same accuracy, we can predict 0.972 hours in advance.

\mname can alleviate the catastrophic forgetting and overfitting. As shown in Figure \ref{fig:result1}a-c, it has the highest BWT and FWT (Section \ref{sec:methods}), meaning it has the lowest negative influence that learning the new tasks has on the old tasks and has the highest positive influence that learning the former data distributions has on the task. Both LM and PM strategies contribute to model performance. As shown in Figure \ref{fig:result1}d-f, if we remove two mechanisms respectively, the model performance will decline.

\begin{figure*}[!t]
\centering
\includegraphics[width=\linewidth]{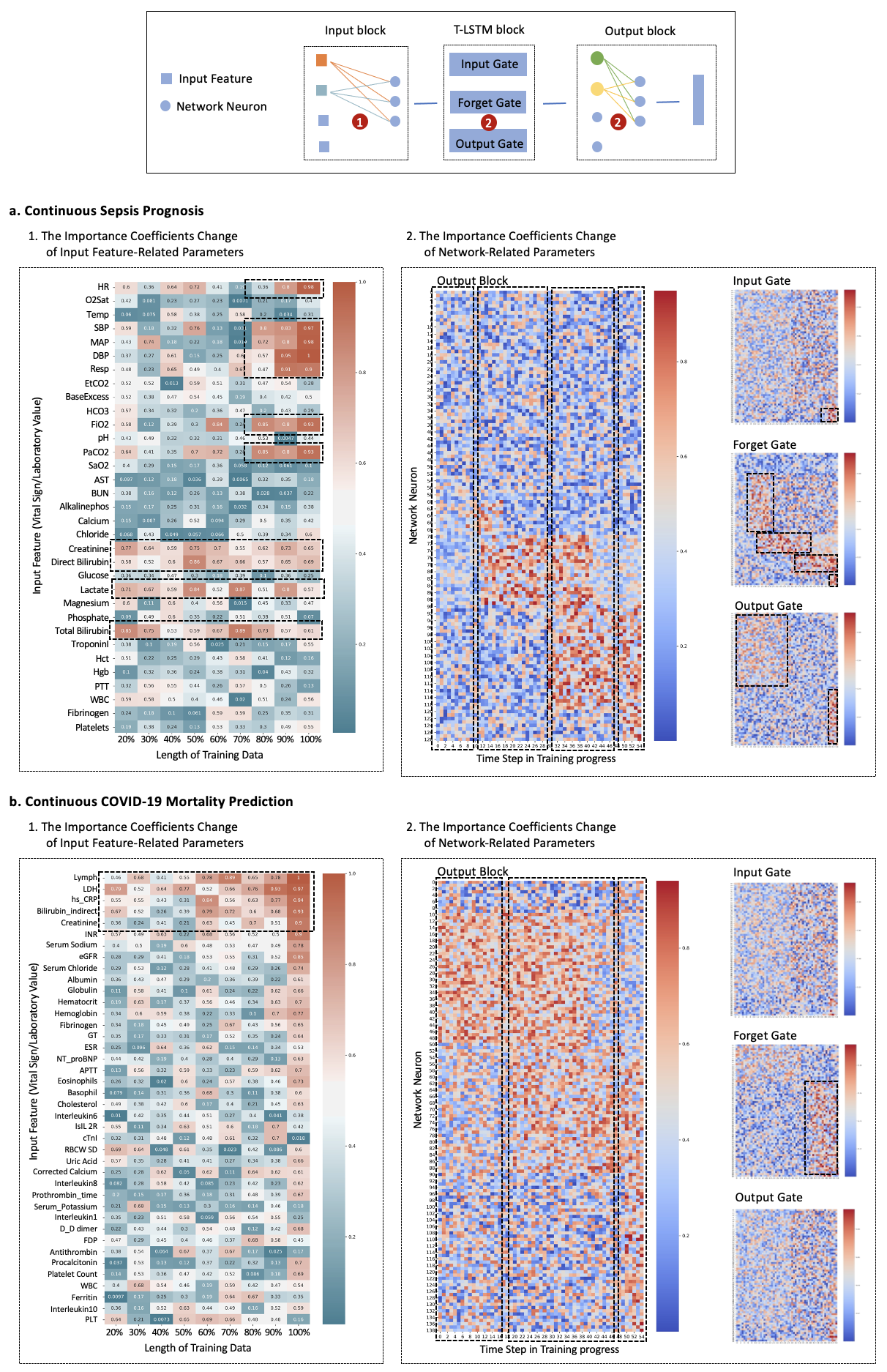}
\caption{The Importance of Input Features and Network Parameters}
\label{fig:result2}
\end{figure*}

\begin{figure*}[!t]
\centering
\includegraphics[width=\linewidth]{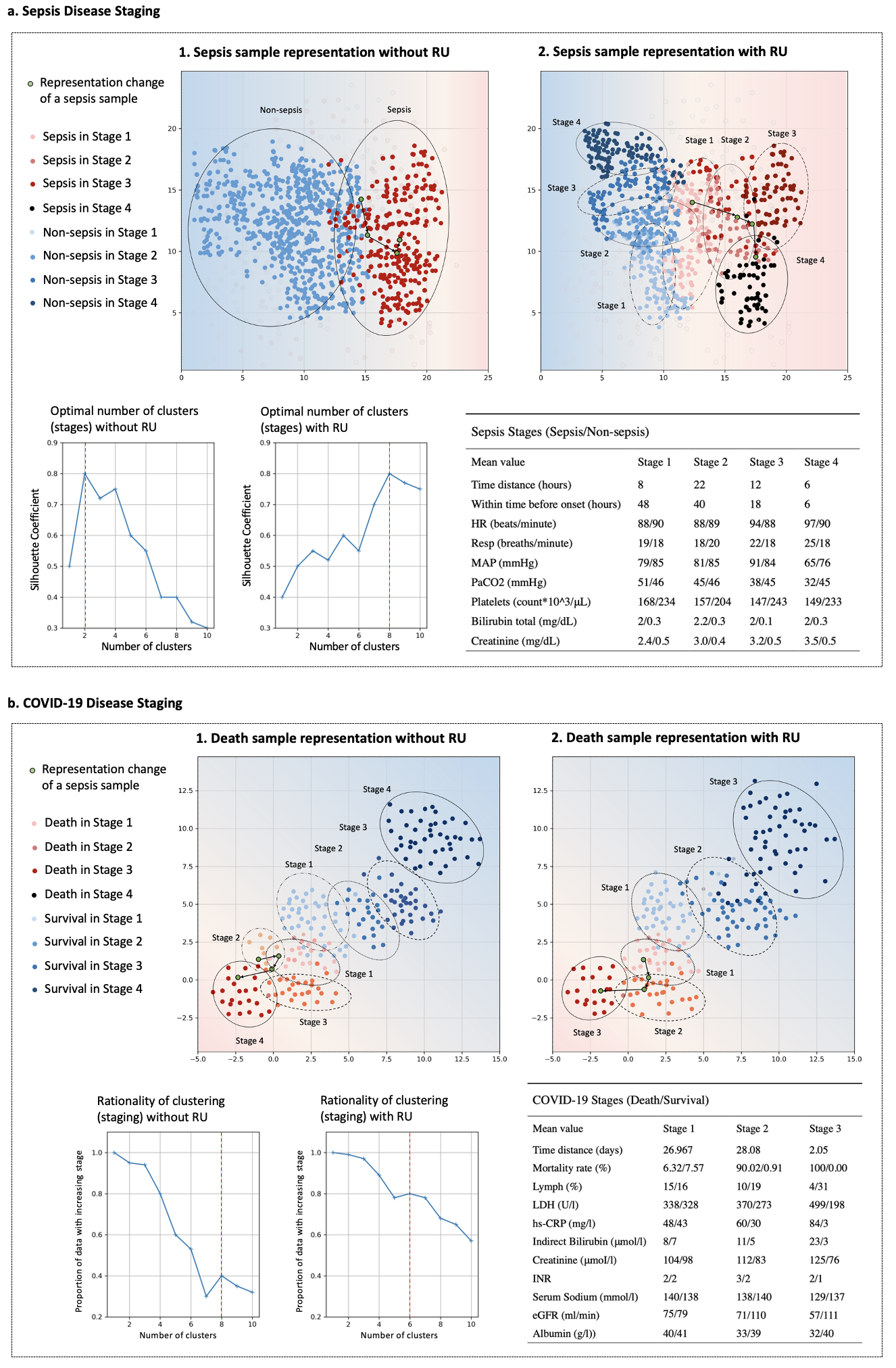}
\caption{Disease Stages and Biomarkers of Sepsis and COVID-19}
\label{fig:result3}
\end{figure*}

\subsubsection*{Finding 2: The change of importance coefficients interprets the learning process of deep learning models.} 

When the model learns time series in different stages, its parameters are updated constantly, and the importance coefficients also change: If the model encounters a new data distribution, the importance coefficient is likely to change significantly. Thus, we explain the learning process of the DL model from the perspective of the change of importance coefficient.

We divided the model into three blocks as shown in Figure \ref{fig:result2}. Block 1 is the input block. We focus on the parameter update process related to input features. For an input feature $x_{i}$, we use the overall importance coefficient of its related parameters to measure its importance: $\alpha^{*}(x_{i})=\sum_{n} \alpha(\theta_{x_{i},l_{n}})$, $\theta_{x_{i},l_{n}}$ is the weight between input feature $x_{i}$ and $n$-th neuron $l_{n}$ in layer $l$; Block 2 is the T-LSTM block. We focus on the parameter update process related to different gates. For a gate $G_{i}$, we use the overall importance coefficient of its parameters to measure its importance: $\alpha^{*}(G_{i})=\sum_{n} \alpha(\theta_{n})$; Block 3 is the output block. We focus on the parameter update process related to network neurons. For $j$-th neuron $l_{i,j}$ in layer $l_{i}$, we use the overall importance coefficient of its output weights to measure its importance: $\alpha^{*}(l_{i,j})=\sum_{n} \alpha(\theta_{l_{i,j},l_{i+1,n}})$. The test on Block 1 aims to enhance the model interpretability from the perspective of data, Block 2 and 3 are about network structure. 

As shown in Figure \ref{fig:result2}a1 and b1, when the model learns time series of different lengths (in different time stages), its perceptual sensitivity to input features is different. For example, for sepsis diagnosis, the importance coefficient $\alpha^{*}$ of the blood pressure increases, which means that the model's perception of blood pressure improved in the later stage; for COVID-19 mortality prediction, the importance coefficient $\alpha^{*}$ of Lymphocytes is always lager, which means that the model pays great attention to this feature at different stages. $\alpha^{*}$ of input features shows their importance. It can be used to evaluate biomarkers.  

As shown in Figure \ref{fig:result2}a2 and b2, in the process of continuous learning, important model parameters are changing. This change process can be divided into several stages. For example, in the output block, the training process of the model can be roughly divided into 4 stages for sepsis diagnosis, and 3 stages for COVID-19 mortality prediction. In each stage, the important parameters are different. In T-LSTM block, this change is obvious for the output gate, but not obvious for the input gate and output gate. These observations reveal the intrinsic mechanism of model learning under our \mname: For different stages of time series (different distributions), the deep learning model activates different neurons to perceive data. This also shows the potential of wide neural networks for \tname. Networks with more neurons in one layer are more likely to learn multi-distributed data.

\subsubsection*{Finding 3: Continuous prognosis reveals the disease biomarkers and stages.}

Semantically, the important feature is the input that has a great impact on the classification results. To quantify them, we define that the important feature is the input with a large overall importance coefficient $\alpha^{*}$. Thus, for tasks of medical diagnosis and prognosis, we can find biomarkers of specific diseases through this method. As shown in Figure \ref{fig:result2}a1 and b1, for sepsis, biomarkers are heart rate (HR), respiration (Resp), mean arterial pressure (MAP), PaCO2, platelets count, total bilirubin and creatinine. For COVID-19, biomarkers are lymphocytes (lymph), lactic dehydrogenase (LDH), high-sensitivity C-reactive protein (hs-CRP), indirect bilirubin, creatinine, etc.

 The response change of the model when the learning disease data continuously can reflect the change of the disease. As shown in Figure \ref{fig:result2}a2 and b2, the training process of the model can be roughly divided into 4 stages for sepsis diagnosis and 3 stages for COVID-19 mortality prediction. Thus, we get the disease stage of these two diseases. By visualizing the hidden layer of block 3 at different stages, we get Figure \ref{fig:result3}. Sepsis has 4 disease stages. Each stage has different reference levels of biomarkers. In some cases, the closer to the onset time, the greater the difference in biomarker reference levels in different prognoses. For example, in stage 1 (the interval from early 48 hours to early 40 hours before the onset time), the respiration difference between sepsis class and non-sepsis is 1, while in stage 4 (the interval from early 6 hours to the onset time), the respiration difference is 7. In other cases, reference levels of biomarkers with different prognoses are different at all stages, such as creatinine. These two conditions may explain the two mechanisms of sepsis. The first is that acute sepsis onset will lead to changes in some vital signs, e.g., a drop in blood pressure, increased lactate, and tachycardia. The second is that patients with congenital characteristics are more likely to get sepsis, e.g., nephropathy with abnormal creatinine and hepatopathy with increased total bilirubin. COVID-19 has 3 disease stages. Compared with sepsis, its data with different classes are more different in the presentation space. This also explains the higher accuracy of continuous COVID-19 mortality prediction than that of continuous sepsis diagnosis. Besides, in the presentation space, the hidden features of the two classes in the later stage are further apart. This shows the difficulty in early classification: the conflict between earliness and accuracy.

\begin{figure*}[!t]

\begin{minipage}[c]{\textwidth} 
\centering
\captionof{table}{COVID-19 Classification Accuracy (AUC-ROC$\uparrow$) with Non-uniform Training Sets and Validation Sets}
\footnotesize
\setlength{\tabcolsep}{0.9mm}{
\begin{tabular}{l|lllllll}
\toprule 
   &SR  &ECEC       &ORGFW    &GEM  &CLOPS    &\mname   \\
\midrule 
Male   &0.968±0.014   &0.969±0.016    &0.965±0.004 &0.978±0.009 &0.978±0.014   &0.971±0.010\\
Female    &0.935±0.004   &0.947±0.015    &0.938±0.003 &0.919±0.008$\downarrow$ &0.921±0.009$\downarrow$   &0.947±0.002\\
\midrule
Age 30-   &0.965±0.014   &0.967±0.015    &0.964±0.009 &0.972±0.008 &0.979±0.012   &0.972±0.010\\
Age 30+  &0.911±0.007$\downarrow$   &0.913±0.018$\downarrow$    &0.923±0.040 &0.932±0.006 &0.914±0.007$\downarrow$   &0.945±0.006\\
\bottomrule 
\end{tabular}}\label{tb:result4_1}\vspace{+0.5cm}
\end{minipage}

\begin{minipage}[c]{\textwidth} 
\centering
\captionof{table}{Classification Accuracy (AUC-ROC$\uparrow$) of \mname with Different Orders of Training Sets}
 \footnotesize
\setlength{\tabcolsep}{1.3mm}{
\begin{tabular}{l|lllllll}
\toprule 
 &Order   &20\%  &40\%      &60\%    &80\%  &100\%      \\
\midrule 
 \multirow{2}*{SEPSIS}  &Time  &0.735±0.003    &0.826±0.003 &0.841±0.003 &0.860±0.005   &0.872±0.001\\
    &Similarity   &0.734±0.006    &0.824±0.004 &0.843±0.004 &0.863±0.007   &0.870±0.002\\
\midrule
 \multirow{2}*{COVID-19}  &Time   &0.789±0.002   &0.902±0.002 &0.926±0.000 &0.959±0.001   &0.968±0.000\\
 &Similarity    &0.790±0.001   &0.910±0.003 &0.924±0.001 &0.958±0.001   &0.967±0.000\\
\bottomrule 
\end{tabular}}\label{tb:result4_2}\vspace{+0.5cm}
\end{minipage}

\begin{minipage}[c]{\textwidth}
\centering
\captionof{table}{Performance (AUC-ROC$\uparrow$, BWT$\uparrow$) for Two Meteorological Datasets.\newline
\footnotesize {UCR-EQ dataset \cite{UCRArchive} has 471 earthquake records from UCR time series database archive. It is the univariate time series of seismic feature value. Natural disaster early warning, like earthquake warning, helps to reduce casualties and property losses \cite{2021Earthquake}. USHCN dataset \cite{USHCN} has the daily meteorological data of 48 states in U.S. from 1887 to 2014. It is the multivariate time series of 5 weather features. Rainfall warning is not only the demand of daily life, but also can help prevent natural disasters \cite{2021The}. }}
 \footnotesize
\setlength{\tabcolsep}{1mm}{
\begin{tabular}{l|lllllll}
\toprule 
   &SR  &ECEC       &ORGFW    &GEM  &CLOPS    &\textbf{\mname}   \\
\midrule 
\multirow{2}*{UCR-EQ}    &0.902±0.002   &0.909±0.010    &0.920±0.001 &0.921±0.001 &0.919±0.004   &\textbf{0.931±0.004}\\
  &0.003 &0.033 &0.112 &0.123 &0.149 &\textbf{0.162}\\
\midrule
\multirow{2}*{USHCN}    &0.911±0.012   &0.902±0.012    &0.9160±0.004 &0.920±0.003 &0.921±0.005   &\textbf{0.930±0.005}\\
  &0.034 &0.047 &0.072 &0.098 &0.082 &\textbf{0.124}\\
\bottomrule 
\end{tabular}} \label{tb:result4_3} \vspace{+0.5cm}
\end{minipage}

\begin{minipage}[c]{\textwidth} 
\centering
\captionof{table}{Performance (AUC-ROC$\uparrow$/BWT$\uparrow$) of Different Neural Networks with \mname}
 \tiny
\setlength{\tabcolsep}{0.6mm}{
\begin{tabular}{l|lllllll}
\toprule 
   &Sepsis  &COVID-19       &10-Diseases    &UCR-EQ  &USHCN      \\
\midrule 
LSTM   &0.867±0.008/0.002 &0.909±0.003/0.047 &0.786±0.002/0.054  &0.881±0.004/0.032 &0.891±0.003/0.054\\
+\mname  &0.907±0.008/0.065 &0.969±0.003/0.115 &0.856±0.002/0.102  &0.931±0.004/0.162 &0.930±0.005/0.124\\
\midrule
CNN   &0.858±0.002/0.004 &0.903±0.002/0.037 &0.774±0.004/0.032  &0.878±0.005/0.030 &0.881±0.004/0.057\\
+\mname  &0.904±0.003/0.067 &0.960±0.006/0.075 &0.849±0.002/0.099  &0.929±0.006/0.150 &0.922±0.005/0.118\\
\midrule 
Transformer   &0.843±0.011/0.005 &0.906±0.005/0.040 &0.784±0.006/0.059  &0.889±0.010/0.029 &0.880±0.015/0.059\\
+\mname  &0.903±0.008/0.067 &0.960±0.007/0.109 &0.852±0.008/0.124  &0.920±0.008/0.132 &0.921±0.008/0.120\\
\bottomrule 
\end{tabular}}\label{tb:result4_4}
\end{minipage}
\end{figure*}

\subsubsection*{Finding 4: Restricted update strategy enhances the ability of the model for atypical scenarios and sustainable use.} 

\mname can avoid model overfitting and guarantee certain model generalization. In Table \ref{tb:result4_1}, we divided the dataset according to gender and age, for most baselines, the accuracy on the validation set is much lower than that on the training set. Mark $\downarrow$ means the accuracy is greatly reduced over 5\%. \mname helps the model maintain robustness. Meanwhile, \mname can prevent the result difference caused by the different orders of training sets. The method we have introduced is to use time series of different stages to train the model, and the order is based on time. Another order is the data similarity \cite{2021A} as many vital signs are periodic. For example, the cycle of blood pressure is one day. Therefore, after using vital signs within 24 hours, we will use the data within 25 hours according to the time order, but use the data within 48 hours according to the similarity order. No matter what order is adopted, \mname has stable accuracy as shown in Table \ref{tb:result4_2}. It shows the potential of PM's global optimization and the potential of \mname's sustainable use.

Furthermore, \mname is a data-agnostic, model-agnostic, and easy-to-use plug-in. It can not only improve the accuracy of continuous classification of medical time series but also play a role in other fields. As shown in Table \ref{tb:result4_3}, \mname outperforms baselines on meteorological data for tasks of continuous earthquake early warning and rainfall prediction: It has the best accuracy of continuous classification and the best ability to resist forgetting. \mname can also be used to train other DL models such as Convolutional Neural Network (CNN) \cite{DBLP:journals/nn/0058S21} and Transformer \cite{vaswani2017attention}. It is easy to use and does not need to change the network structure. As shown in Table \ref{tb:result4_4}, if we use \mname to train base models, the accuracy can be improved by more than 5\%. And \mname is not limited by hyper-parameters. The hyper-parameters are $\rho$ and $\lambda$. $\rho$ determines the correlation between current and previous gradients in Equation \ref{eq:d_m}. We find that PM performs well when $\rho$ is the same as the learning rate $\rho_{t}=\eta_{t}=\frac{1}{(t+1)^{a}}, a=1$. $\lambda$ decides the constraint degree on parameter update in Equation \ref{eq:LP}. We can optimize it using the search method supplied by mature tools.

\section*{Discussion}\label{sec:discussion}

\textbf{Deep learning model has the potential to explore disease mechanisms}. The importance coefficient not only explained the working mechanism of the DL model but also dug out the disease biomarkers and stages. Different from the statistics and case analysis of the medical gold standard, these biomarkers are based on the judgment basis of the DL model. It can provide a new horizon for medical research. For example, based on the learning process of the DL model, for sepsis, a drop in blood pressure, increased lactate, and tachycardia are important in the later stage, while abnormal creatinine and total bilirubin are always important. It can be explained that sepsis is an acute disease and is related to some congenital diseases like nephropathy and hepatopathy. Such behavior is in exact accordance with the sepsis literature \cite{2022renal,2017Liver}; For COVID-19, only lymph, LDH, and hs-CRP are most important throughout the stages. This shows that COVID-19 has a clear reference to measure the disease severity \cite{2022redifine}. Meanwhile, we match the disease stage with the changes in model parameters during the learning process. In this way, the disease stage is no longer defined only by the level of biomarkers or by clustering and patient subtyping, but by the characteristic changes in the high-dimensional space created by the model. 

\textbf{\mname strategy helps to indefinably disease staging.} At present, except for cancer, it is difficult to define clear stages for most diseases. For sepsis, disease stratification is implemented by recommended clinical criteria (e.g. SIRS \cite{Roger1992Definitions}, SOFA \cite{1996The}, qSOFA \cite{singer2016third}, and etc.). But they focus on severity, not the progression. We emphasize that disease staging is the disease change over time. \mname can achieve this according to the model change when learning the medical time series from different time stages. As shown in Figure \ref{fig:result3}a, \mname can identify stages directly according to the importance coefficient change of model parameters, instead of using unsupervised clustering methods. Without \mname, the clustering method is difficult to find stages: The number of clusters with the best silhouette coefficient is 2, and the number of stages is 1. For COVID-19, most work categorizes it roughly into early stage and late stage \cite{2021Metabolomic}. Some existing DL-based methods can perform disease staging by using representation learning. For example, our previous work \cite{DBLP:journals/BMC/sun} clustered features in hidden layers of T-LSTM and got 4 COVID-19 stages. As shown in Figure \ref{fig:result3}b, this clustering-based method can get a good silhouette coefficient, but can not guarantee the time constraint: When identifying 4 stages, only about 40\% of the samples will be divided into stages corresponding to chronological order. For example, a death sample (green dot) was initially judged to be stage 2, then stage 1, and finally stage 4. But stages 1-4 are in time order. Using \mname, this inconsistency is largely alleviated: In the case of 1-10 clusters, the percentage of samples with a time-increasing stage is raised. The death sample (green dot) was judged as stage 1, then stage 2, and finally stage 3 over time.

\textbf{Learning multi-distributed data is the general trend.} Currently, many sophisticated DL models (e.g., RNN \cite{DBLP:journals/tai/GuptaGBD20}, CNN \cite{DBLP:journals/nn/0058S21}, and Transformer \cite{vaswani2017attention}) have shown outstanding achievements in time series modeling in many fields. For offline learning, after the model has learned the dataset, the model is only sensitive to the learned distribution. For example, when the model has learned the full-length vital signs of sepsis, it usually classifies accurately at the onset time. But it's too late for critical illness. To gain treatment time, the model needs to learn early data. However, there are also problems in learning early data at only one stage. For example, a time series may have missed the learned stage at the beginning, the characteristics of early data are not obvious, requiring late data assistance, etc. Thus, it is necessary for the model to learn time series from different stages, i.e. multi-distributed data. In this way, the model can realize continuous classification of time series.

\textbf{Reasonable model training strategy is the icing on the cake.} Many studies have shown that a meaningful learning strategy can improve the model performance and generalization power \cite{DBLP:conf/naacl/DevlinCLT19}. Therefore, we realize the ability of model learning multi-distributed data from the perspective of model updating strategy. The experimental results show that a reasonable model training strategy plays a key role in improving the model performance. Different from the design of the model structure, the design and application of strategy will be more extensive. It pays more attention to the overall goal and has few requirements for specific data and used models. That is, our method is data-agnostic, model-agnostic, and easy-to-use.

\textbf{Quantifying the updating process makes it possible to interpret the deep learning model}. Interpretability remains one of the key issues to be solved to achieve the trust of clinicians and insert the deep learning algorithm into clinical workflow. DL models are often considered to be black-box because they typically have high-dimensional nonlinear operations, many model parameters and complex model architectures, which makes them difficult for a human to understand. Therefore, showing the parameter changes during training has the potential to explain the DL model. In this work, our method \mname implements \tname by updating model parameters with constraints. The constraint is achieved by quantifying the importance of the parameters. Surprisingly, the importance of parameters can be used to explain the deep learning model. As shown in experimental results, it can identify both the input features and the structure part that is important for classification.

\textbf{Opportunities of \tname.} Currently, some sub-disciplines (e.g. online learning, continual learning anomaly detection) also study the mode of continuous classification. But their setting methods can’t satisfy Requirements 1, 2, 3, 4 simultaneously (Appendix \ref{sec:related work}). \tname is a new potential field facing practical problems. Meanwhile, we found that the neural network structure with reasonable width is more conducive to continuous classification and continual learning. Because with the learning of new data distribution, the change of important parameters on the scale of network width is more obvious and regular, but that on the scale of depth is confused. Therefore, future work can study the impact of model structure on \tname from the perspective of network depth and network width. Further, we found that different learning orders have little effect on our method \mname. This demonstrates the potential of our approach for off-line continual learning: We use the existing data to train the model and put it into use. After a period of time, new data may be generated. We can continue to train the current model with new data instead of designing a new model. In addition, the method for \tname can be context-independent in the future. The model can perform not only different medical tasks, but also tasks in other fields like meteorology at the same time (Appendix \ref{sec:experiments}).

\section*{Methods}\label{sec:methods}

\subsection*{Problem Formulation} \label{sec:problem formulation}

\begin{definition} [Continuous Classification, CC] \label{def:CC}
A time series $X=\{x_{1},...x_{M}\}$ is labeled with a class $C\in \mathcal{C}$ at the final time $T$. CC classifies $X$ at every $t$ with loss $\sum_{t=1}^{M}\mathcal{L}(f(X_{1:m}),C)$.
\end{definition}

Without the loss of generality, we use the univariate time series to present the problem. Multivariate time series can be described by changing $x_{m}$ to $x_{m,i}$. $i$ is the i-th dimension. Note that single-shot classification optimizes the objective with a single loss $\mathcal{L}(f(x),c)$. \mname should consider the multi-distribution and classify more times. 

\begin{definition} [Continuous Classification of Time Series ]\label{def:CCTS} A dataset $\mathcal{T}=\{X^{n}\}_{n=1}^{N}$ contains $N$ time series. Each time series $X^{n}=\{x^{n}_{m}\}_{m=1}^{M}$ has $M$ observations with value $x^{n}_{m}$ at time $t^{n}_{m}$. At the final time $t^{n}_{M}$, $X^{n}$ is labeled with a class $C\in \mathcal{C}$. As time series varies among time, it has a subsequence series with $M$ different distributions $\mathcal{D}=\{\mathcal{D}_{m}\}_{m=1}^{M}$. \mname learns every $\mathcal{D}_{m}$ and introduces a task sequence $\mathcal{M}=\{\mathcal{M}^{m}\}_{m=1}^{M}$ to minimize the additive risk $\sum_{m=1}^{M}\mathbb{E}_{\mathcal{M}^{m}}[\mathcal{L} (f^{m}(\mathcal{D}_{m};\theta),C)]$ with model $f$ and parameter $\theta$. $f^{m}$ is the model $f$ after being trained for $\mathcal{M}^{m}$. When the model is trained for $\mathcal{M}^{m}$, its performance on all observed data cannot degrade: $\frac{1}{m}\sum_{i=1}^{m}\mathcal{L}(f^{i},\mathcal{M}^{i}) \leq \frac{1}{m-1}\sum_{i=1}^{m-1} \mathcal{L}(f^{i},\mathcal{M}^{i})$.
\end{definition}

\subsection*{Time-aware Long Short-Term Memory}\label{sec:t-lstm}

Recurrent neural networks (RNNs) take sequence data as input, recursion occurs in the direction of sequence evolution, and all units are chained together. In classical RNN, the current state $h_{t}$ is affected by the previous state $h_{t-1}$ and the current input $x_{t}$ and is described as $h_t=\sigma(Wx_{t}+Uh_{t-1}+b)$, where $\sigma$ is an activation function, and $W$, $U$ and $b$ are learnable parameters. 

However, the real-world time series, especially vital signs, have long sequences and are irregularly sampled. The classical RNN only process uniformly distributed longitudinal data by assuming that the sequences have an equal distribution of time differences. Thus, we implement Time-aware Long Short-Term Memory (T-LSTM) \cite{Baytas:2017:PSV:3097983.3097997} to solve the long-term dependency problem and capture the irregular temporal dynamics. Based on the classical LSTM, T-LSTM possesses some new designs. $C_{m-1}^{S}$ component learns the short-term memory of sequence by learnable network parameters. $C_{m-1}^{T}$ is the long-term memory calculated from the former memory cell $C_{m-1}$  with getting rid of $C_{m-1}^{S}$. $C_{m-1}^{S}$ is adjusted to the discounted short-term memory $\hat{C}_{m-1}^{S}$ by the elapsed time function $g(\Delta_{t})$. The previous memory $C_{m-1}^{*}$ is changed to the complement subspace of $C_{m-1}^T$ combined with $\hat{C}_{m-1}^S$. 

\begin{equation}
    \begin{aligned}
        C_{m-1}^{S}=\tanh(W_{d}C_{m-1}+b_{d})\ \ \ \ \ \ \ \ \ \ \ \ \ \ \text{Short-term memory}&\\ 
        \hat{C}_{m-1}^S=C_{m-1}^S\cdot g(\Delta_{t}) \ \ \ \ \ \ \ \ \text{Discounted short-term memory}&\\
        C_{m-1}^{T}=C_{m-1}-C_{m-1}^{S}  \ \ \ \ \ \ \ \ \ \ \ \ \ \ \ \ \ \  \ \ \ \    \text{Long-term memory}&\\
        C_{m-1}^{*}=C_{m-1}^{T}-\hat{C}_{m-1}^{S}  \ \ \ \ \ \ \ \ \ \ \ \  \text{Adjusted previous memory}&\\
        f_{m}=\sigma(W_{f} x_{m}+U_{f} h_{m-1}+b_{f})  \ \ \ \ \ \ \ \ \ \ \ \ \ \ \ \ \ \ \ \ \ \text{Forget gate}&\\
        i_{m}=\sigma(W_{i} x_{m}+U_{i} h_{m-1}+b_{i}) \ \ \ \ \ \ \ \ \ \ \ \ \ \ \ \ \ \ \ \ \ \ \ \ \text{Input gate}&\\
        \tilde{C}_{m}=\tanh(W_{c} x_{m}+U_{c} h_{m-1}+b_{o})\ \ \ \  \quad\text{Candidate memory}&\\
        C_{m}=f_{m}\cdot C_{m-1}^{*}+i_{m}\cdot \tilde{C}_{m}  \ \ \ \ \ \ \ \ \ \ \ \ \ \ \ \ \ \ \ \ \text{Current memory}&\\
        o_{m}=\sigma(W_{o} x_{m}+U_{o} h_{m-1}+b_{o})  \ \ \ \ \ \ \ \ \ \ \ \ \ \ \ \ \ \ \ \ \text{Output gate}&\\
        h_{m}=o_{m} \cdot \tanh(C_{m}) \ \ \ \ \ \ \ \ \ \ \ \ \ \ \ \ \ \ \ \ \ \ \ \text{Current hidden state}&
    \end{aligned}
\end{equation}

We use a log calculation for the elapsed time function. $\Delta_{t}$ describes the time gap between two records at two sequential time steps $t_{m}$ and $t_{m-1}$.

\begin{equation}
    \begin{aligned}
        g(\Delta_{t})&=\frac{1}{\log (e+\Delta_t)}  \\
        \Delta_{t}&=t_{m}-t_{m-1}
    \end{aligned}
\end{equation}

\subsection*{Restricted Update Strategy for Neural Network Parameters}\label{sec:ru}

\subsubsection*{Limitation Mechanism} \label{sec:LM}

When the model meets a distribution, it will change from $f_{m-1}$ to $f_{m}$. In order to let the model performance on all tasks not degrade, the loss $\mathcal{L}$ of the current $f^{m}$ on tasks $\{\mathcal{M}^{k}\}_{k=1}^{m}$ should be not bigger than that of the previous $f^{m-1}$ on tasks $\{\mathcal{M}^{k}\}_{k=1}^{m-1}$:

\begin{equation} \label{eq:CLgoal}
\begin{aligned}
    &\underset{\theta^{m}}{\min}\ \mathcal{L}(f^{m}(X_{1:m},\theta^{m}),C)\\
    \text{subject to }\frac{1}{m}\sum_{k=1}^{m}&\mathcal{L}(f^{m},\mathcal{M}^{k}) \leq \frac{1}{m-1}\sum_{k=1}^{m-1} \mathcal{L}(f^{m-1},\mathcal{M}^{k})
\end{aligned}
\end{equation}

The fundamental cause of catastrophic forgetting is that the arbitrary change of neural network parameters leads to calculation errors on old tasks. Based on the observation of many parameter configurations resulting in the same performance \cite{DBLP:journals/nn/ParisiKPKW19}, we could add a regular term to the loss to restrict the updating of model parameters. Thus, in LM, we constrain important parameters to stay close to their old values, but change unimportant parameters more. We give a new loss in Equation \ref{eq:CLloss}. A regularization term is added to the original loss $\mathcal{L}$, where $\alpha$ is the importance coefficient of parameter $\theta$. With the minimum $\mathcal{O}$, $\theta^{m}$ will be changed less from $\theta^{m-1}$ with a large $\alpha$. As shown in Figure \ref{fig:method}, $\theta_{\text{important}}$ is limited in a region.

\begin{equation} \label{eq:CLloss}
\mathcal{O}(\theta^{m})=\mathcal{L}(f^{m}(\theta^{m}),\mathcal{M}^{m})+\lambda\sum_{i}\alpha_{i}(\theta^{m}_{i}-\theta^{m-1}_{i})^{2}
\end{equation}

The second derivative of probability can evaluate the importance coefficient $\alpha=(\log\mathrm{p}(D_{m}\vert\theta^{m}))''$. Elastic weight consolidation \cite{DBLP:journals/corr/KirkpatrickPRVD16} defines $\alpha$ from a probabilistic perspective $\log \mathrm{p}(\theta\vert\mathcal{D})=\log \mathrm{p}(D_{m}\vert\theta)+\log \mathrm{p}(\theta\vert D_{m-1})-\log \mathrm{p}(D_{m})$: Optimizing the parameters is tantamount to finding their most probable values under $\mathcal{D}$. The posterior probability is indicated by Laplace approximation. 

Taking this inspiration, we use the diagonal of Fisher information matrix \cite{DBLP:journals/corr/abs-1301-3584} to represent the first-order derivatives.

\begin{equation} \label{eq:F_}
\rm{F}=\frac{1}{N}\sum_{k=1}^{N}\nabla \log p(D_{k}\vert\theta)\nabla \log p(D_{k}\vert\theta)^{\top}
\end{equation}

Thus, we represent the importance coefficient $\alpha$ by $F$ and re-arrange Equation \ref{eq:CLloss} to:

\begin{equation} \label{eq:LP}
\mathcal{O}(\theta^{m})=\mathcal{L}(f^{m}(\theta^{m}),\mathcal{M}^{m})+\lambda\sum_{i}\rm{F_{i}}(\theta^{m}_{i}-\theta^{m-1}_{i})^{2}
\end{equation}

\begin{equation} \label{eq:F}
\rm{F_{i}}=\frac{1}{m}\sum_{k=1}^{m}(\frac{\partial\log p(D^{k}\vert \theta^{m}_{i})}{\partial\theta^{m}_{i}})^{2}
\end{equation}

\subsubsection*{Promotion Mechanism} \label{sec:PM}

When we focus on the final task, if the model meets a new distribution, the learned knowledge will be part of the final solution. We regard this as a continuous optimization problem. In this way, different data distributions are treated equally. The new data helps the model learn the old data, which can reduce the unstable solution caused by the different learning orders. Continuous optimization problem \cite{DBLP:journals/corr/abs-1909-05207} is defined as regret minimization. For task $\mathcal{M}$, the regret $\mathcal{R}$ is the difference between the total loss and that of the best parameter $\theta^{*}$ of the fixed decision in hindsight.
\begin{equation} \label{eq:regret}
\small
\mathcal{R}_{M}:=\sum_{m=1}^{M}(\mathcal{L}(f^{m}(\theta^{m}),\mathcal{M})-\mathcal{L}(f^{m}(\theta^{m*}),\mathcal{M}))
\end{equation}

For regret minimization, we design a Promotion Mechanism (PM) with mechanisms of projection-free and stochastic recursive gradient. It focuses the quality of the final performance instead of iterates produced from the course of optimization. For continuous optimization, the main bottleneck is the computation of projections onto the underlying decision set $\underset{\mathcal{K}}{\prod}(\theta)=\underset{\hat{\theta}\in\mathcal{K}}{\arg \min}\vert\vert\hat{\theta}-\theta\vert\vert$ \cite{DBLP:conf/aaai/WangLH020}. The projection operation is defined as the closest point inside the convex set $\mathcal{K}$ of Euclidean space to a given point. Projection-free methods, like Frank-Wolfe \cite{DBLP:conf/icml/ChenHHK18}, can replace the projection with a linear optimization at each iteration. It alleviates the complexity but remains problems of training non-converging and instability \cite{DBLP:journals/corr/abs-2010-12493}. 

Thus, we estimate a stochastic recursive estimator based on stochastic gradient technology \cite{DBLP:conf/nips/CutkoskyO19,DBLP:conf/aaai/XieSZWQ20}. Assuming for task $\mathcal{M}^{m}$, the model receives new time series data $X_{1:m}$ with distribution $D_{m}$ and gets the loss $\mathcal{L}$. We first give a random variable $\xi_{m}$ satisfying:

\begin{equation}
    \mathbb{E}_{\xi_{t}\sim D_{m}}[\nabla L(\theta^{m},\xi_{m})]=\nabla \sum_{m=1}^{M} \mathcal{L}(f^{m}(X_{1:m},\theta^{m}),C)
\end{equation}

Then the stochastic recursive estimator is:

\begin{equation} \label{eq:stochastic recursive estimator}
\rm{d_{m}}=\nabla \mathcal{L}(\theta^{m},\xi_{m})+(1-\rho_{t})(\rm{d_{m-1}}-\nabla \mathcal{L}(\theta^{m-1},\xi_{m}))
\end{equation}

It finds a solution $\rm{v_{m}}$ of the linear optimization problem

\begin{equation} \label{eq:PP1}
\rm{v_{m}}=\arg\min_{\rm{v}\in\mathcal{K}}\vert\vert\rm{d_{m}},\rm{v_{m}}\vert\vert_{2}
\end{equation}

to update $\theta^{m+1}$ in the direction of gradient $\rm{g_{m+1}}$:

\begin{equation} \label{eq:PP2}
  \rm{g_{m+1}}=\rm{v_{m}}-\theta^{m}, \quad
  \theta^{m+1}=\theta^{m}+\eta_{m}\cdot \rm{g_{m+1}}
\end{equation}

Such method randomly selects samples to guide the change of gradient and leads to faster converges. It could achieve a nearly optimal ${O}(\sqrt{M})$ regret bound with high probability.

\subsubsection*{Overall Training Process}

Gradient Episodic Memory (GEM) \cite{DBLP:conf/nips/Lopez-PazR17} shows that when the new gradient $\rm{g}_{m}$ and the old gradients $\rm{g}_{k}$ are at an acute angle, the model performance on the old task will improve, at least not decrease: 

\begin{equation} \label{eq:GEM}
    \left <\rm{g}_{m},\rm{g}_{k}\right>=\left<\frac{\partial \mathcal{L}(f^{m},\mathcal{M}^{m})}{\partial \theta^{m}},\frac{\partial \mathcal{L}(f^{k},\mathcal{M}^{k})}{\partial \theta^{k}}\right>\geq 0,k=1,...,m-1
\end{equation}

Thus, the regularization is projecting the current gradient $\rm{g}_{m}$ to the closest gradient $\rm{g}_{m}'$ by satisfying all the constraint of acute angle:

\begin{equation} \label{eq:GEM2}
    \min\vert\vert\rm{g_{m}}-\rm{g_{m}'}\vert\vert_{2},\textit{ }\text{subject to  }\left <\rm{g_{k}},\rm{g_{m}'}\right>\geq 0  \textit{, } k=1,...,m-1
\end{equation}

In LM, $\rm{F}$ is positive semi-definite \cite{DBLP:conf/nips/SmolaVE03}. This property not only guarantees that seeing each task as a factor of the posterior (LM) but also guarantees the acute angle change of a vector after the product (PM). Thus, \mname updates network parameters by using the regularized loss $\mathcal{O}$ in Equation \ref{eq:LP} instead of the basic loss $\mathcal{L}$. And we re-arrange Equation \ref{eq:stochastic recursive estimator} to

\begin{equation} \label{eq:d_m}
    \rm{d_{m}}\leftarrow\nabla_{\theta^{m}}\mathcal{O}^{m}+(1-\rho_{m})(\rm{d_{m-1}}-\rm{g^{m-1}})
\end{equation}

\subsubsection*{Regret and Complexity}

PM holds a nearly optimal regret bound $\tilde{O}(\sqrt{M})$. W.p. at least $1-\delta$ for any $\delta \in(0,1)$, $\mathcal{R}_{M}\leq (logT+1)(f(\theta^1)-f(\theta^{*}))+(16LD^2+16\sigma+4B)\sqrt{2Mlog\frac{8M}{\delta}}+\frac{1}{2}LD^2(logM+1)^2$. Where $\eta_{m},\rho_{m}=\frac{1}{m+1}$, $D$ is diameter of convex set, $L$ is $L$-Lipschitz-continuous. PM and LM achieve $O(1)$ per-round computational cost. If the complexity of training a base model to convergence is $O$ and data length is $M$, the overall complexity will be $MO$. More details of mathematical derivation are in Appendix \ref{sec:mathematics}.

\subsubsection*{Evaluation Metrics}

The classification results are evaluated by assessing the area under the curve of the Receiver Operating Characteristic (AUC-ROC). The ROC is a curve of the True Positive Rate (TPR) and the False Positive Rate (FPR). TN, TP, FP and FN represent true positives, true negatives, false positives and false negatives, respectively.

\begin{equation} \label{eq:AUC-ROC}
TPR=\frac{TP}{TP+FN}, FPR=\frac{FP}{TN+FP}
\end{equation}

The continuous classification performance are evaluated by Backward Transfer (BWT) and Forward Transfer (FWT). They are the influence that learning a task has on old and future tasks. $R_{i,j}$ is the accuracy of task $\mathcal{M}^{j}$ after completing task $\mathcal{M}^{i}$. $\overline{b}$ is the accuracy with random initialization. 

\begin{equation} \label{eq:BWT}
\text{BWT}=\frac{1}{\vert\mathcal{M}\vert-1}\sum_{i=1}^{\vert\mathcal{M}\vert-1}R_{\vert\mathcal{M}\vert,i}-R_{i,i}
\end{equation}

\begin{equation} \label{eq:FWT}
\text{FWT}=\frac{1}{\vert\mathcal{M}\vert-1}\sum_{i=2}^{\vert\mathcal{M}\vert}R_{i-1,i}-\overline{b}_{i,i}
\end{equation}

\subsection*{Data availability}
All data used in this paper are publicly available and can be accessed as follows: SEPSIS \cite{DBLP:conf/cinc/ReynaJSJSWSNC19}, COVID-19 \cite{COVID-19}, MIMIC-III \cite{johnson2016mimic}, USHCN \cite{USHCN}, UCR \cite{UCRArchive}.

\subsection*{Code availability}
The code is available at \url{https://github.com/SCXsunchenxi/CCTS}.

\bibliography{sn-bibliography}

\section*{Acknowledgments}
This work was supported by the National Natural Science Foundation of China (No.62172018, No.62102008), and the National Key Research and Development Program of China under Grant 2021YFE0205300.

\section*{Author contributions}
C.S. and H.L. conceived the project. C.S. and S.H contributed ideas, designed and conducted the experiments. S.H, H.L., M.S, D.C., B,Z evaluated the experiments. All authors co-wrote the manuscript.

\section*{Competing interests}
The authors declare no competing interests.

\section*{Additional information}
\backmatter
\bmhead{Supplementary information}
See Appendix.
\bmhead{Correspondence and requests for materials}
should be addressed to Chenxi Sun, Hongyan Li, and Shenda Hong.

\clearpage

\begin{appendices}

\section{Related Work and Concepts} \label{sec:related work}

Time series is one of the most common data forms, the popularity of time series classification has attracted increasing attention in many practical fields, such as healthcare and industry. In the real world, many applications require classification at every time. For example, in the Intensive Care Unit (ICU), critical patients' vital signs develop dynamically, the status perception and disease diagnosis are needed at any time. Timely diagnosis provides more opportunities to rescue lives. In response to the current demand, we propose a new task -- Continuous Classification of Time Series (\tname). It aims to classify as accurately as possible at every time in time series.

\begin{figure*}[!h]
\centering
\includegraphics[width=\linewidth]{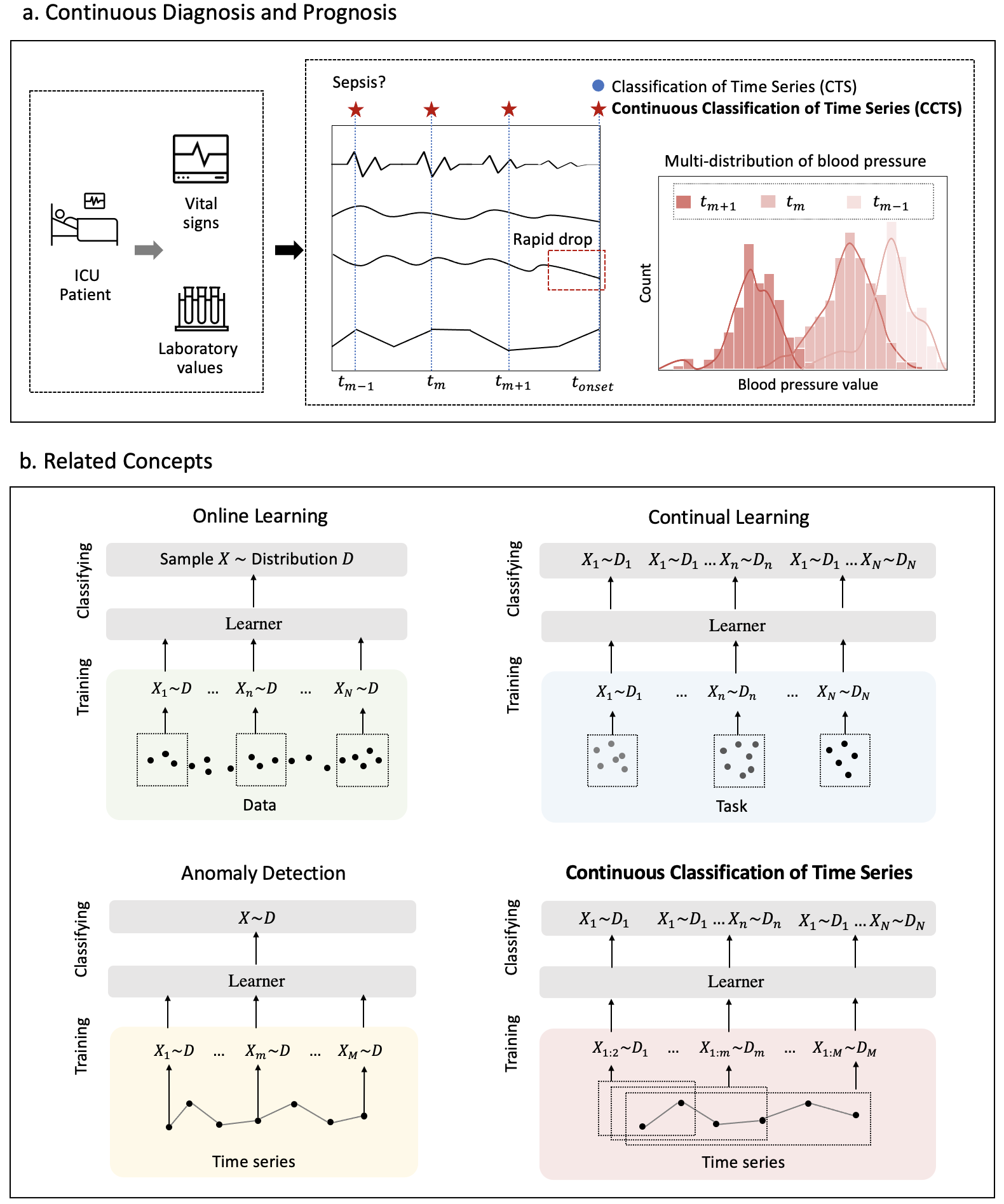}
\caption{Continuous Classification of Time Series (\tname) \newline Differences and Similarities between \tname and Other Concepts}
\label{fig:concept}
\end{figure*}

Currently, some sub-disciplines also study the mode of continuous learning or continuous classification. But their setting does not match our needs and their methods can't address our issues. As shown in Figure \ref{fig:concept}, Online Learning (OL) \cite{DBLP:conf/aaai/WanXZ21} models the incoming data steam continuously to solve an overall optimization problem with the partially observed data. It focuses more on issues in data steam, rather than the dynamics of time series. OL cannot meet the Requirement 1, 2, 4; Continual Learning (CL) \cite{9349197} enables the model to learn new tasks over time without forgetting the old tasks. In its setting, the model learns a new task at every moment. The old task and new task are clear so that the multi-distribution is fixed. While the dynamic time series has data correlation over time, which easily further causes the overfitting problem. CL cannot meet the Requirement 2 and partial Requirement 1; Anomaly Detection (AD) \cite{DBLP:journals/csur/FernandoGDSF22} identifies data that does not conform to the expected pattern. It mainly maintains one data distribution and gives an alarm when an exception occurs. AD cannot meet Requirement 1 and partial Requirement 2. Because the existing research can not meet the current demand, we propose a new task \tname.

The existing work can be summarized into two categories.

\subsection{Single-shot Classification}

Classifying at a fixed time. \textit{ A time series $X=\{x_{1},...x_{T}\}$ is labeled with classes $C$. Single-shot classification aims to classify $X$ at a time $t, t\leq T$ with the minimum loss $\mathcal{L}(f(X_{1:t}),C)$.}

The foundation is the Classification of Time Series (CTS), making classification based on the full-length data \cite{DBLP:journals/datamine/FawazFWIM19}. But in time-sensitive applications, Early Classification of Time Series (ECTS), classifying at an early time, is more critical  \cite{DBLP:journals/tai/GuptaGBD20}. For example, early diagnosis helps for sepsis outcomes \cite{DBLP:conf/aaai/LiuLSGN18}. Nowadays, Recurrent Neural Networks (RNNs) and Convolutional Neural Networks (CNNs) have shown good performances for CTS and ECTS by modeling long-term dependencies \cite{choi2017using}, addressing data irregularities \cite{ijcai2021-414}, learning frequency features \cite{DBLP:conf/pakdd/HsuLT19}, etc.

\begin{definition} [Classification of Time Series (CTS)] \label{def:CTS}
A dataset of time series $\mathcal{D}=\{(X^{n},C^{n})\}_{n=1}^{N}$ has $N$ samples. Each time series $X^{n}$ is labeled with a class $C^{n}$, CTS classifies time series using the full-length data by model $f:f(X)\to C$
\end{definition}

\begin{definition} [Early Classification of Time Series (ECTS)] \label{def:ECTS}
A dataset of time series $\mathcal{D}=\{(X^{n},C^{n})\}_{n=1}^{N}$ has $N$ samples. Each time series $X^{n}={X_{t}}_{t=1}^{T}$ is labeled with a class $C^{n}$. ECTS classifies time series in an advanced time $t$ by model $f:f(\{x_{1},x_{2},...,x_{t}\})\to C$, where $t<T$.
\end{definition}

The existing (early) classification of time series is the single-shot classification, where the classification is performed only once at the final or an early time. However, many real-world applications require continuous classification. For example, intensive care patients should be detected and diagnosed at all times to facilitate timely life-saving. The above methods only classify once and just lean a single data distribution. They have good performances on i.i.d data at a fixed time, like early 6 hours sepsis diagnosis \cite{2019Early}, but fail for multi-distribution. In fact, continuous classification is composed of multiple single-shot classifications as shown in Figure \ref{fig:introduction}.

\subsection{Continuous Classification}

Classifying at every time. \textit{A time series is $X=\{x_{1},...x_{T}\}$. At time $t$, $x_{1:t}$ is labeled with class $c_{t}$. Continuous Classification classifies $x_{1:t}$ at every time $t=1,...,T$ with the minimum loss $\sum_{t=1}^{T}\mathcal{L}(f(x_{1:t}),c^{t})$.}

Most methods use multi-model to learn multi-distribution, like SR \cite{DBLP:journals/tnn/MoriMDL18} and ECEC \cite{lv2019effective}. They divide data by time stages and design different classifiers for different distributions. But the operation of data division and classifier selection will cause additional losses.

In fact, CCTS is composed of multiple ECTS and the continuous classification is composed of multiple single-shot classification.

\begin{definition} [Continuous Classification of Time Series (\tname)] \label{def:CCTS_}
A dataset of time series $\mathcal{D}=\{(X^{n},C^{n})\}_{n=1}^{N}$ has $N$ samples. Each time series $X^{n}={X_{t}}_{t=1}^{T}$ is labeled with a class $C^{n}$. CCTS classifies time series in every time $t$ by model  $f:f(\{x_{1},x_{2},...,x_{t}\})\to C$, where $t=1,...,T$.
\end{definition}

Currently, some sub-disciplines also study the mode of continuous learning or continuous classification. But their setting does not match our needs and their methods can't address our issues. As shown in Figure \ref{fig:introduction}, Online Learning (OL) \cite{DBLP:conf/aaai/WanXZ21} models the incoming data steam continuously to solve an overall optimization problem with the partially observed data. It focuses more on issues in data steam, rather than the dynamics of time series. Thus, OL cannot meet the Requirement 1, 2, 3; Continual Learning (CL) \cite{9349197} enables the model to learn new tasks over time without forgetting the old tasks. In its setting, the model learns a new task at every moment. The old task and new task are clear so that the multi-distribution is fixed. While the dynamic time series has data correlation over time, which easily further causes the overfitting problem. Thus, CL cannot meet the Requirement 2 and partial Requirement 1; Anomaly Detection (AD) \cite{DBLP:journals/csur/FernandoGDSF22} identifies data that does not conform to the expected pattern. It mainly maintains one data distribution and gives an alarm when an exception occurs. Thus, AD cannot meet Requirement 1 and partial Requirement 2. Because the existing research can not meet the current demand, we propose a new concept \tname.

\begin{definition} [Online Learning (OL)] \label{def:OL}
A OL issue has a sequence of dataset $\mathcal{X}=\{X^{1},X^{2},...,X^{N}\}$ for one task $\mathcal{T}$. Each dataset $X^{t}$ has a distribution $D^{t}$. CL learns a new $D^{t}$ at every time $t$. The goal is to find the optimal solution of $\mathcal{T}$ after $N$ iterations by minimize the regret $\mathcal{R}:=\sum_{t=1}^{N}(f^{t}(X^{t})-\min f^{t}(X^{t}))$.
\end{definition}

\begin{definition} [Continual Learning (CL)] \label{def:CL}
A CL issue $\mathcal{T}=\{T^{1},T^{2},...,T^{N}\}$ has a sequence of $N$ tasks. Each task $T^{n}=(X^{n},C^{n})$ is represented by the training sample $X^{n}$ with classes $C^{n}$. CL learns a new task at every moment. The goal is to control the statistical risk of all seen tasks $\sum_{n=1}^{N}\mathbb{E}_{(X^{n},C^{n})}[\mathcal{L} (f_{n}((X^{n};\theta),C^{n})]$ with loss $\mathcal{L}$, network function $f_{n}$ and parameters $\theta$.
\end{definition}

\section{Mathematics} \label{sec:mathematics}

\begin{assumption}
The compact convex set $\mathcal{C}\subseteq \mathbb{R}^{d}$ has diameter $D$. $\forall \theta_{1},\theta_{2} \in \mathcal{C}$,
\begin{equation}
    \vert\vert\theta_{1}-\theta_{2}\vert\vert\leq D
\end{equation}
\end{assumption}

\begin{assumption}
The stochastic gradient $\nabla F_{t}(\theta,\xi_{t})$ is unbiased with $\mathbb{E}_{\xi_{t}}[\nabla F_{t}(\theta,\xi_{t})]=\nabla f_{t}(\theta)$ and is $L$-Lipschitz continuous over the constraint set $\mathcal{C}$ with
\begin{equation}
    \vert\vert\nabla F_{t}(\theta_{1},\xi_{t})=\nabla F_{t}(\theta_{2,\xi_{t}})\vert\vert\leq L\vert\vert\theta_{1}=\theta_{2}\vert\vert, \forall \theta_{1},\theta_{2}\in \mathcal{C}.
\end{equation}
\end{assumption}

The above Assumption immediately implies that $f_{t}$ is differentiable and has $L$-Lipschitz-continuous gradients.

In the stochastic online setting, we denote the expected loss function as the $\overline{f}=\mathbb{E}_{f_{t}\thicksim \mathcal{D}[f_{t}]}$. In order to obtain high probability results. We assume the following:

\begin{assumption}
The distance between the stochastic gradient $\nabla F_{t}(\theta,\xi_{t})$ and the exact gradient is bounded over the constraint set $\mathcal{C}$, for any $theta \in \mathcal{C}, t\in \{1,...,T\}$, there exist $\sigma < \infty$ such that with probability $1$,
\begin{equation}
    \vert\vert\nabla F_{t}(\theta,\xi_{t})-\nabla\overline{f}(\theta)\vert\vert^{2}\leq \sigma^{2}
\end{equation}
The difference of $f_{t}(\theta)$ and $\overline{f}_{t}(\theta)$ is bounded over the constraint set $\mathcal{C}$. $\forall \theta \in \mathcal{C}, t\in \{1,...,T\}$, there exist $M^{2}<\infty$ such that with probability $1$,
\begin{equation}
    \vert f(\theta)-\overline{f}(\theta)\vert\leq M^{2}
\end{equation}
\end{assumption}

We show that the norm of the gradient estimation error $\varepsilon_{t}:=d_{t}-\nabla\overline{f}(\theta_{t})$ converges to zero rapidly w.h.p. 

First, we reformulate $\varepsilon_{t}$ as the sum of a martingale difference sequence $\{\epsilon_{t,k}\}_{k=1}^{t}$ w.r.t. a filtration $\{\mathcal{F}_{t}\}_{k=0}^{t}$, i.e., $\varepsilon_{t}=\sum_{k=1}^{t}\epsilon_{t,k}$, where $\mathbb{E}[\epsilon_{t,k}\vert\mathcal{F}_{t-1}\vert]=0$ and $\mathcal{F}_{t-1}$ is the $\theta$-filed generate by $\{f_{i},\xi_{i}\}_{i=1}^{k-1}$. By showing that $\vert\vert\epsilon_{t,k}\vert\vert\leq c_{t,k}$ for some constant $c_{t,k}$, one can relate the Hoeffding-type concentration inequality. With carefully chosen $\{\rho_{t}\}_{t=1}^{T}$ and $\{\eta_{t}\}_{t=1}^{T}$, the quantity $q_{t}$ can be shown to converge to $0$ at a sublinear rate by induction. As a result, $\vert\vert\varepsilon_{t}\vert\vert$ converges to zero at a sublinear rate w.h.p. as stated in the following lemma. $D$ is diameter of convex set, $L$ is $L$-Lipschitz-continuous.

\begin{lemma}
With $\rho_{t}=\eta_{t}=\frac{1}{(1+t)^{a}}$ for some $a\in (,1]$, if Assumptions are satisfied for any $t\geq 1$ and $\delta_{0}\in (0,1)$ we have w.p.at least $1-\delta_{0}$,
\begin{equation}
    \vert\vert\varepsilon_{t}\vert\vert\leq 2(2LD)+\frac{3^{a}\sigma}{3^{a}-1}(t+1)^{-\frac{a}{2}}\sqrt{2\log(\frac{4}{\delta_{0}})}.
\end{equation}
\end{lemma}

Lemma 1 shows that the gradient approximation error $\vert\vert\varepsilon_{t}\vert\vert$ converges to zero at a fast sublinear rate $\tilde{\mathcal{O}}(\frac{1}{t^{\frac{a}{2}}})$ w.h.p if $\rho_{t}=\eta_{t}=\frac{1}{(1+t)^{a}}$ for any $a\in (0,1]$. This result is critical to the regret analysis of our methods.

\begin{theorem}
With $\rho_{t}=\eta_{t}=\frac{1}{1+t}$. If $\overline{f}$ is convex and Assumption are satisfied, then w.p. at least $1-\delta$ for any $\delta \in(0,1)$ for any $\delta \in(0,1)$,
\begin{equation}
    \mathcal{R}_{T}\leq (logT+1)(f(\theta^1)-f(\theta^{*}))+(16LD^2+16\sigma+4B)\sqrt{2Tlog\frac{8T}{\delta}}+\frac{1}{2}LD^2(logT+1)^2
\end{equation}
\end{theorem}

\section{Experiments} \label{sec:experiments}

\subsection{Datasets}

We use 6 datasets to test methods.

 SEPSIS dataset \cite{DBLP:conf/cinc/ReynaJSJSWSNC19} has 30,336 records with 2,359 diagnosed sepsis. Early diagnose is critical to improve sepsis outcome \cite{seymour2017time}. In this dataset, the time series are the changes of 40 related patient features, the label at each time is sepsis or non-sepsis. Early diagnose can improve sepsis outcome.
 
 COVID-19 dataset \cite{COVID-19} has 6,877 blood samples of 485 COVID-19 patients from Tongji Hospital, Wuhan, China. Mortality prediction helps for treatment and rational resource allocation \cite{DBLP:journals/BMC/sun}. In this dataset, the time series are the changes of blood samples, the label at each time is mortality or survival. Mortality prediction helps for personalized treatment and rational resource allocation
 
 MIMIC-III dataset \cite{johnson2016mimic} has 19,993 admission records of 7,537 patients. We focus on 8 diagnoses (ICD-9): Diabetes(249), Hypertension (401), Heart Failure (428), Pneumonia (480-486), Gastric Ulcer (531), Hepatopathy (571), Nephropathy (580-589), SIRS (995.9). The time series are vital signs, and labels at each time are some diagnoses.
 
 USHCNrain dataset \cite{USHCN} has the daily meteorological data of 48 states in U.S. from 1887 to 2014. It is the multivariate time series of 5 weather features. Rainfall warning is not only the demand of daily life, but also can help prevent natural disasters.
 
 USHCN dataset \cite{USHCN} has U.S. daily meteorological data from 1887 to 2014. We focus on 4 weather conditions in New York: sunny, overcast, rainfall, snowfall. The time series are records of 4 neighboring states, labels at each time are weather after a weak.
 
 UCR-EQ dataset has 471 earthquake records from UCR time series database archive. It is the univariate time series of seismic feature value. Natural disaster early warning, like earthquake warning, helps to reduce casualties and property losses.

\subsection{Baselines}

We use 9 related work as baselines.

ECTS-based methods: LSTM \cite{choi2017using} trains a model by time series at every time stage; SR \cite{DBLP:journals/tnn/MoriMDL18} gives the fusion result of multiple models trained by the full-length data; ECEC \cite{lv2019effective} has trains a set of classifiers by data in different time stages.
 
CL-based methods: EWC \cite{DBLP:journals/corr/KirkpatrickPRVD16} is a regularization-based method, training a model to remember the old tasks by constraining important parameters to stay close to their old values. GEM \cite{DBLP:conf/nips/Lopez-PazR17} update parameters by finding the gradients which are at acute angles to the old gradients; CLEAR is a replay-based method, using the reservoir sampling to limit the number of stored samples to a fixed budget assuming an i.i.d. data stream; CLOPS \cite{2021A} replays old tasks to avoid forgetting. 

OL-based methods: OSFW \cite{DBLP:conf/icml/ChenHHK18} uses stochastic gradient estimator; ORGFW \cite{DBLP:conf/aaai/XieSZWQ20} uses recursive gradient estimator.

\subsection{Settings}

The classification results are evaluated by assessing the area under the curve of the Receiver Operating Characteristic (AUC-ROC). The ROC is a curve of the True Positive Rate (TPR) and the False Positive Rate (FPR). TN, TP, FP and FN represent true positives, true negatives, false positives and false negatives, respectively.

\begin{equation} \label{eq:AUC-ROC_}
TPR=\frac{TP}{TP+FN}, FPR=\frac{FP}{TN+FP}
\end{equation}

The continuous classification performance are evaluated by Backward Transfer (BWT) and Forward Transfer (FWT). They are the influence that learning a task has on old and future tasks. $R_{i,j}$ is the accuracy of task $\mathcal{M}^{j}$ after completing task $\mathcal{M}^{i}$. $\overline{b}$ is the accuracy with random initialization. 

\begin{equation} \label{eq:BWT_}
\text{BWT}=\frac{1}{\vert\mathcal{M}\vert-1}\sum_{i=1}^{\vert\mathcal{M}\vert-1}R_{\vert\mathcal{M}\vert,i}-R_{i,i}
\end{equation}

\begin{equation} \label{eq:FWT_}
\text{FWT}=\frac{1}{\vert\mathcal{M}\vert-1}\sum_{i=2}^{\vert\mathcal{M}\vert}R_{i-1,i}-\overline{b}_{i,i}
\end{equation}

Learning stability evaluation is fluctuation:

\begin{equation} \label{eq:r}
R = \frac{1}{n-1}\sqrt{\sum_{i=1}^{n}(d_{i}-d_{i-1})^{2}}. 
\end{equation}

\begin{equation} \label{eq:r_} 
d=\left\{
    \begin{aligned}
        -1, \text{if }g<0\\
        1, \text{if }g>0
    \end{aligned}
\right.
\end{equation}

All methods use the same LSTM as base model. All datasets are divided into training, test, validation of 6:2:2. Results are got by 5-fold cross-validation. The code is available at \url{https://github.com/PaperCodeAnonymous/CCTS}.

\subsection{Results and Analysis}

\noindent \textbf{Multiple Distribution}

\begin{figure*}[!h]
\caption{Model Gradient Fluctuation}
\centering
\centerline{
\includegraphics[width=0.7\linewidth]{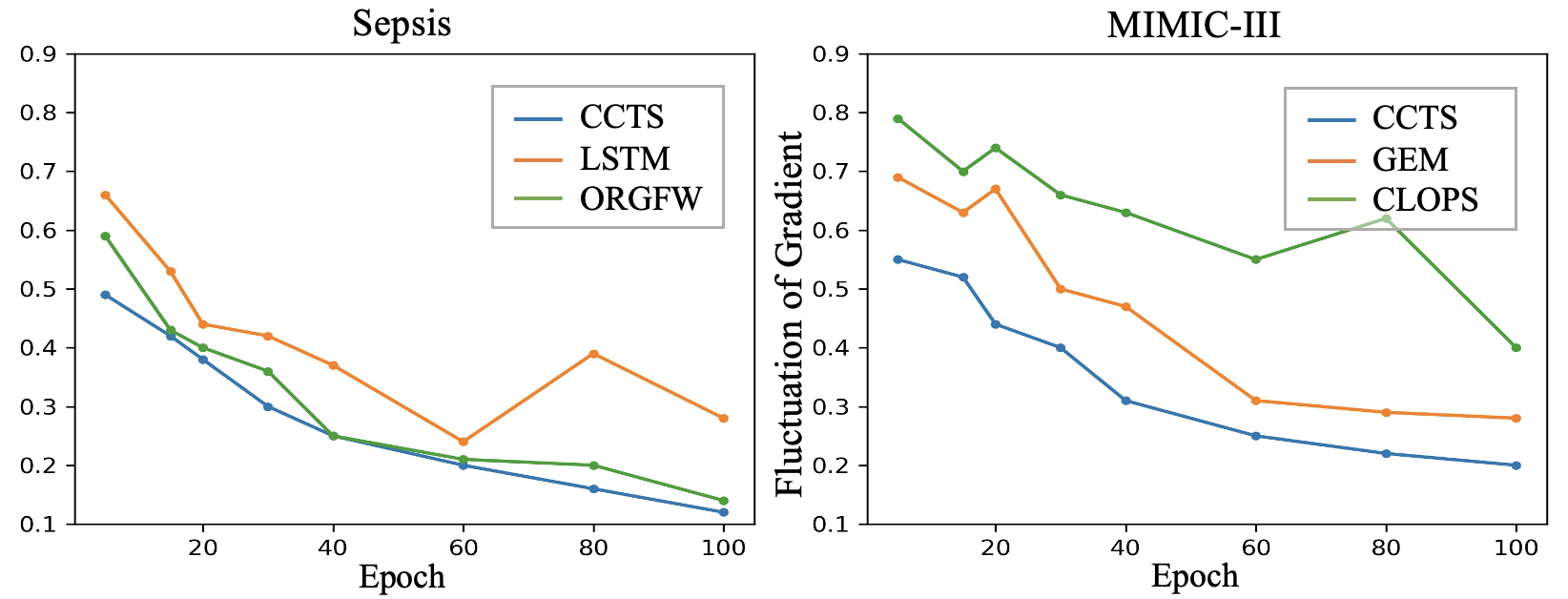}}
\label{fig:result_1}
\end{figure*}

Before discussing the method performance, we show the basic scenario of CCTS -- multi-distribution in Figure \ref{fig:result_1}. The data in different time stages have distinct statistical characteristics and finally form multiple distributions. The fundamental goal of the following experiment is to model them.

\noindent \textbf{Result of classification accuracy}

Our method is significantly better than baselines. In Bonferroni-Dunn test, $k=6$, $n=4$, $m=5$ are the number of methods, datasets, cross-validation fold, then $N=n\times m=20, \text{CD}=1.524$, finally $\text{rank(RU)}=1<\text{CD}+1$. Thus, the accuracy is significantly improved. As shown in Table \ref{tb:accuracy}, \mname can classify more accurately at every time. The average accuracy is about 2\% higher, especially in the early time, being 5\% higher for 10\%-length data. Take sepsis diagnosis as an example, compared with the best baseline, our method improves the accuracy by 1.4\% on average, 2.2\% in the early 50\% time stage when the key features are unobvious. Each hour of delayed treatment increases sepsis mortality by 4-8\% \cite{seymour2017time}. With the same accuracy, we can predict 0.972 h in advance.

\noindent \textbf{Analysis of continuous classification}

\begin{figure*}[!h]
\caption{Classification Accuracy Change, Model Gradient Fluctuation}
\centering
\centerline{
\includegraphics[width=1\linewidth]{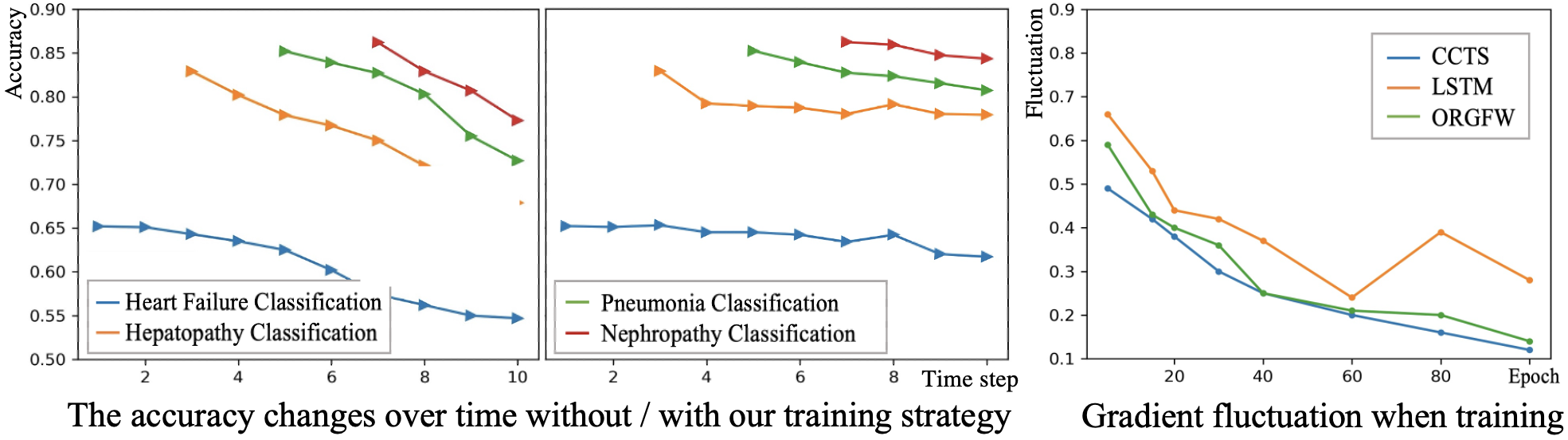}}
\label{fig:result}
\end{figure*}

 Our strategy can alleviate the catastrophic forgetting and promote the overall performance by sub-distribution. As shown in Table \ref{tb:accuracy}, \mname has the best performance on the early time series, showing the ability of LM to alleviate catastrophic forgetting. As shown in Table \ref{tb:BWT} and \ref{tb:FWT}, \mname has the highest BWT, meaning it has the lowest negative influence that learning the new tasks has on the old tasks. Figure \ref{fig:result} shows the case study of 4 tasks about how \mname overcome the accuracy degradation. As shown in Table \ref{tb:BWT} and \ref{tb:FWT}, \mname has the highest FWT, meaning it has the highest positive influence that learning the former data distributions has on the task, especially for Sepsis and COVID-19 datasets. In Table \ref{tb:overfitting}, for most baselines, the accuracy on validation set is much lower than that on training set. Mark $\downarrow$ means the accuracy is greatly reduced over 5\%.

\noindent \textbf{Ablation study}

 Both LM and PM strategies contribute to model performance. As shown in Table \ref{tb:ablation}, if we remove two strategies respectively, the model accuracy will decline, the relation between tasks will become worse, the model instability will increase. Besides, our method is a training strategy, except RNNs, it can also be applied to other DL models, e.g., CNN, TCN.

\noindent \textbf{Analysis of gradient stability} 

 Our strategy has the most stable gradients in training process. As shown in Figure \ref{fig:result}, CCTS has the smallest value of $R$ in any training epoch, It shows the restriction ability of our method in error back propagation of DNNs.

\noindent \textbf{Analysis of class number and training order} 

\begin{figure}[!h]
\caption{Task Similarity \newline
\small{In MIMIC-III dataset, the diagnoses with ICD-9 order are 1:HIV, 2:Brain Cancer, 3:Diabetes, 4:Hypertension, 5:Heart Failure, 6:Pneumonia, 7:Gastric Ulcer, 8:Hepatopathy, 9:Nephropathy, 10:SIRS. The new similarity order is 1, 10, 2, 4, 5, 8, 3, 9, 7, 6.}}
\centering
\includegraphics[width=0.5\linewidth]{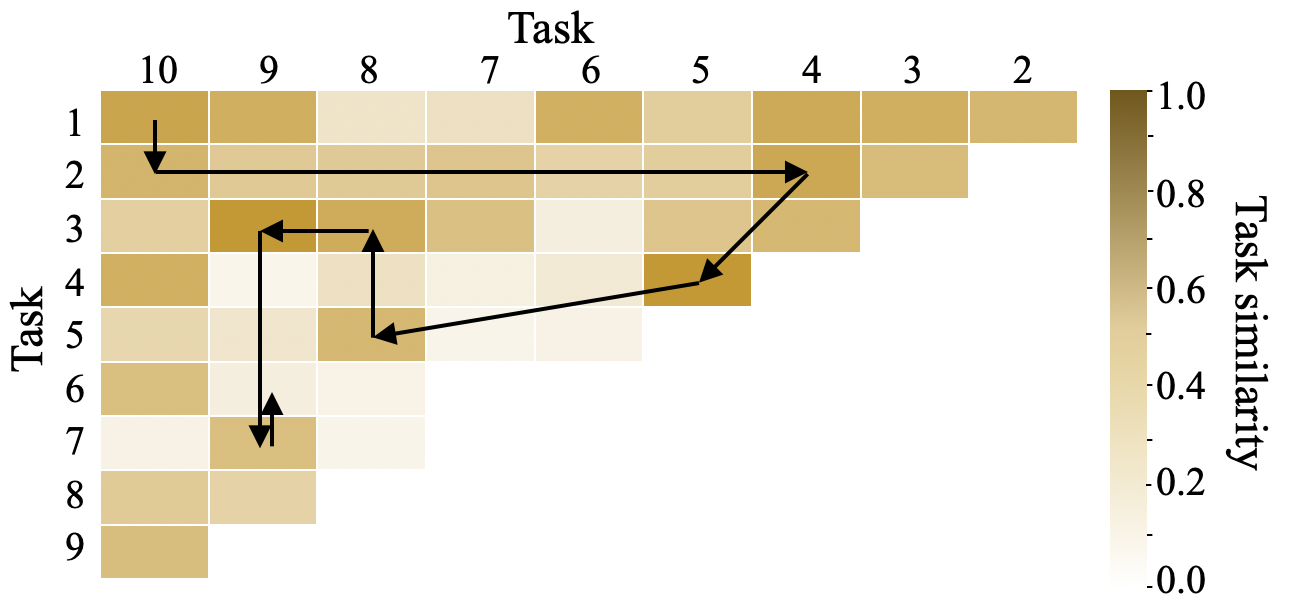}
\label{fig:appendix3}
\end{figure}

 Class number and training order will influence the result: 
Fewer classes lead to better performance of \mname; A sound training order can improve the model performance. As shown in Table \ref{tb:tasknumber}, if we increase the diagnosis number in MIMIC-III, the accuracy will decrease. It's also common in other methods \cite{DBLP:journals/nn/ParisiKPKW19}. Becides, No matter what order is adopted, RU has stable accuracy. It shows the possibility of global optimization potential of PM in \mname. According to the Gaussian distribution $\mathcal{N}(\mu,\sigma^{2})$ in Figure \ref{fig:introduction} and the task similarity of task $i$ and task $j$ in curriculum learning literature, we can obtain a new task order in Figure \ref{fig:appendix3}. 

\begin{equation}
S(i,j)=1-\sqrt{1-\sqrt{\frac{2\sigma_{i}\sigma_{j}}{\sigma{i}^{2}+\sigma{j}^{2}}e^{-\frac{1}{4}\frac{(\mu_{i}-\mu_{j})^{2}}{\sigma{i}^{2}+\sigma{j}^{2}}}}}
\end{equation}

\noindent \textbf{Hyper-parameter setting} 

 The hyper-parameters in PM and LM are $\rho$ and $\lambda$. $\rho$ determines the correlation between current and previous gradients in Equation \ref{eq:d_m}. We find that PP performs well when $\rho$ is the same as the learning rate $\rho_{m}=\eta_{m}=\frac{1}{(m+1)^{a}}, a=1$. $\lambda$ decides the constraint degree on parameter update in Equation \ref{eq:LP}. We optimize it using the search method supplied by mature tools.

\clearpage

\begin{table*}[!ht] 
\caption{Classification Accuracy (AUC-ROC$\uparrow$) for 4 Real-world Datasets at the First 5 Time Steps.\newline
\small {*k\% means the current classification time is k\% of the total time of the full-length time series; Bold font indicates the highest accuracy.}} \label{tb:appendix2}
\centering
\footnotesize
\setlength{\tabcolsep}{0.7mm}{
\begin{tabular}{llcccccccccc}
\toprule 
 Dataset &Method &10\%    &20\%   &30\%    &40\%    &50\%  \\
\midrule 
\multirow{8}*{UCR-EQ} 
& LSTM &0.695±0.044    &0.711±0.038   &0.803±0.024    &0.843±0.019    &0.854±0.017 \\
& SR &0.700±0.015    &0.736±0.014   &0.830±0.016    &0.863±0.015   &0.871±0.024 \\
& ECEC &0.703±0.013  &0.738±0.018   &0.828±0.017    &0.865±0.014   &0.873±0.026   \\
& EWC &0.724±0.015  &0.768±0.018   &0.848±0.014    &0.874±0.016   &0.883±0.025    \\
& GEM &0.723±0.014    &0.767±0.017    &0.850±0.015   &0.876±0.016   &0.890±0.024\\
& CLEAR &0.729±0.015   &0.770±0.015   &0.852±0.019   &0.880±0.013  &0.899±0.026\\
& CLOPS &0.728±0.016   &0.773±0.016   &0.855±0.015  &0.878±0.016   &0.896±0.028   \\
&\textbf{\mname}  &\textbf{0.730±0.022} &\textbf{0.774±0.023}    &\textbf{0.856±0.015}    &\textbf{0.882±0.022}    &\textbf{0.900±.0017}       \\

\midrule 
\multirow{8}*{USHCNrain}
& LSTM &0.682±0.014  &0.700±0.028   &0.721±0.013  &0.745±0.028    &0.784±0.023   \\
& SR &0.702±0.014 &0.730±0.022   &0.745±0.016    &0.761±0.023    &0.809±0.024  \\
& ECEC &0.707±0.017  &0.736±0.024   &0.748±0.015    &0.760±0.025   &0.806±0.025 \\
& EWC &0.727±0.018  &0.736±0.025   &0.768±0.017    &0.798±0.024    &0.805±0.022  \\
& GEM &0.720±0.019  &0.728±0.026    &0.772±0.015   &0.781±0.023    &0.801±0.026    \\
& CLEAR &0.728±0.016  &0.738±0.025    &0.773±0.018  &0.784±0.024   &0.802±0.027 \\
& CLOPS &0.728±0.012  &0.740±0.024   &0.769±0.019  &0.781±0.025  &0.800±0.024 \\
&\textbf{\mname} &\textbf{0.730±0.018} &\textbf{0.742±0.017}    &\textbf{0.775±0.016}    &\textbf{0.791±0.021}   &\textbf{0.810±.0133}    \\

\midrule 
\multirow{8}*{COVID-19\ } 
& LSTM &0.605±0.044   &0.701±0.033    &0.793±0.022   &0.833±0.015    &0.844±0.013  \\
& SR &0.636±0.014   &0.730±0.024     &0.810±0.013   &0.867±0.016    &0.901±0.013   \\
& ECEC &0.639±0.013   &0.732±0.028   &0.829±0.013   &0.870±0.016   &0.901±0.026  \\
& EWC &0.703±0.022   &0.769±0.015    &0.870±0.014    &0.888±0.028   &0.915±0.017   \\
& GEM &0.699±0.025    &0.779±0.017    &0.871±0.015    &0.885±0.022   &0.914±0.019  \\
& CLEAR &0.710±0.013    &0.785±0.019   &0.870±0.016   &0.879±0.016   &0.916±0.024  \\
& CLOPS &0.709±0.017    &0.775±0.013    &0.869±0.012   &0.900±0.017  &0.918±0.026  \\
&\textbf{\mname}  &\textbf{0.712±0.021} &\textbf{0.790±0.023}    &\textbf{0.872±0.013}    &\textbf{0.901±0.022}    &\textbf{0.919±0.016}       \\

\midrule 
\multirow{8}*{SEPSIS} 
& LSTM &0.576±0.063   &0.629±0.035    &0.735±0.064    &0.736±0.064   &0.745±0.056 \\
& SR &0.626±0.035    &0.659±0.015    &0.768±0.013    &0.791±0.026    &0.803±0.018  \\
& ECEC &0.623±0.024   &0.669±0.019   &0.761±0.016    &0.793±0.016    &0.811±0.015  \\
& EWC &0.671±0.027   &0.733±0.023    &0.799±0.015    &0.827±0.036    &0.832±0.028  \\
& GEM &0.670±0.026   &0.730±0.024    &0.802±0.018    &0.826±0.033    &0.834±0.026  \\
& CLEAR &0.680±0.028 &0.732±0.024   &0.801±0.015    &0.825±0.035    &0.833±0.025  \\
& CLOPS &0.684±0.025  &0.733±0.025  &0.802±0.017   &0.824±0.036    &0.830±0.023  \\
&\textbf{\mname}  &\textbf{0.690±0.032} &\textbf{0.734±0.038}    &\textbf{0.812±0.022}    &\textbf{0.828±0.036}    &\textbf{0.835±0.024}     \\
\bottomrule 
\end{tabular}
}\label{tb:accuracy}
\end{table*}

\begin{table*}[!h]
\caption{Classification Accuracy (AUC-ROC$\uparrow$) for 4 Real-world Datasets at the Last 5 Time Steps.\newline
\small {*k\% means the current classification time is k\% of the total time of the full-length time series; Bold font indicates the highest accuracy.}}\label{tb:appendix3}
\centering
\footnotesize
\setlength{\tabcolsep}{0.7mm}{
\begin{tabular}{llcccccccccc}
\toprule 
 Dataset &Method   &60\%   &70\%    &80\%    &90\%    &100\%  \\
\midrule 
\multirow{8}*{UCR-EQ} 
& LSTM  &0.874±0.012   &0.913±0.034    &0.909±0.014  &0.919±0.008    &0.924±0.012 \\
& SR &0.888±0.017    &0.924±0.010  &0.928±0.105   &0.936±0.103   &0.941±0.104\\
& ECEC  &0.890±0.015  &0.923±0.013    &0.929±0.107   &0.936±0.006 &0.940±0.009  \\
& EWC  &0.895±0.014   &0.910±0.017    &0.923±0.102 &0.930±0.005     &0.933±0.003\\
& GEM &0.900±0.015   &0.920±0.015   &0.929±0.008  &0.935±0.003     &0.934±0.004 \\
& CLEAR &0.904±0.012   &0.918±0.019   &0.923±0.004    &0.928±0.007     &0.932±0.005\\
& CLOPS  &0.902±0.015  &0.915±0.010   &0.917±0.006   &0.921±0.009    &0.925±0.005 \\
&\textbf{\mname}  &\textbf{0.906±0.005}     &\textbf{0.928±0.007}    &\textbf{0.933±0.010}    &\textbf{0.940±0.005}    &\textbf{0.946±0.003}   \\

\midrule 
\multirow{8}*{USHCNrain}
& LSTM   &0.820±0.015   &0.837±0.024   &0.852±0.014    &0.869±0.025  &0.891±0.002 \\
& SR    &0.836±0.016     &0.886±0.023  &0.902±0.013   &0.921±0.026  &0.933±0.009\\
& ECEC  &0.837±0.016     &0.887±0.027  &0.906±0.017   &0.920±0.028    &0.931±0.009\\
& EWC  &0.834±0.016     &0.867±0.026  &0.896±0.017   &0.906±0.020    &0.926±0.007\\
& GEM    &0.838±0.013    &0.868±0.029   &0.899±0.010   &0.910±0.021    &0.928±0.005\\
& CLEAR   &0.837±0.010  &0.867±0.023 &0.879±0.012 &0.899±0.027    &0.921±0.004\\
& CLOPS    &0.835±0.016   &0.861±0.024  &0.877±0.011   &0.895±0.016    &0.919±0.013\\
&\textbf{\mname}    &\textbf{0.841±0.012}     &\textbf{0.898±0.022}    &\textbf{0.910±0.015}    &\textbf{0.928±0.013}   &\textbf{0.939±0.013}\\

\midrule 
\multirow{8}*{COVID-19\ } 
& LSTM  &0.888±0.013   &0.918±0.033    &0.925±0.014   &0.939±0.005    &0.944±0.015 \\
& SR   &0.900±0.018    &0.935±0.010   &0.946±0.006   &0.952±0.017   &0.962±0.005 \\
& ECEC   &0.904±0.014    &0.937±0.008&0.948±0.015    &0.952±0.008  &0.963±0.017 \\
& EWC    &0.923±0.014    &0.935±0.007   &0.940±0.013   &0.950±0.013   &0.954±0.008\\
& GEM    &0.924±0.018    &0.936±0.009   &0.939±0.010   &0.949±0.017    &0.953±0.005\\
& CLEAR   &0.926±0.014  &0.933±0.011  &0.941±0.007    &0.948±0.009  &0.952±0.008\\
& CLOPS &0.925±0.015    &0.935±0.013   &0.940±0.007  &0.947±0.006  &0.954±0.006\\
&\textbf{\mname}    &\textbf{0.927±0.006}     &\textbf{0.955±0.008} &\textbf{0.960±0.011}    &\textbf{0.963±0.009}    &\textbf{0.967±0.008}   \\

\midrule 
\multirow{8}*{SEPSIS} 
& LSTM  &0.748±0.043    &0.773±0.032    &0.795±0.027    &0.813±0.025   &0.827±0.039\\
& SR &0.827±0.037    &0.835±0.013    &0.845±0.014   &0.859±0.022    &0.866±0.023\\
& ECEC  &0.815±0.014   &0.827±0.016    &0.849±0.016   &0.859±0.017   &0.863±0.014\\
& EWC  &0.838±0.024    &0.842±0.030    &0.848±0.017   &0.850±0.014    &0.854±0.016\\
& GEM  &0.836±0.028    &0.841±0.034   &0.849±0.014   &0.851±0.016    &0.853±0.012  \\
& CLEAR &0.839±0.028   &0.842±0.031   &0.847±0.010  &0.850±0.019   &0.848±0.016 \\
& CLOPS &0.838±0.026   &0.842±0.030   &0.850±0.017 &0.853±0.010   &0.857±0.018 \\
&\textbf{\mname}    &\textbf{0.842±0.034}     &\textbf{0.852±0.023}    &\textbf{0.857±0.012}    &\textbf{0.866±0.014}    & \textbf{0.872±0.012}  \\
\bottomrule 
\end{tabular}
}
\end{table*}

\begin{table}[!h]
\centering
\caption{CL Performance (BWT$\uparrow$) of Baselines. \newline \small{$^{1}$}LSTM, SR and ECEC are not listed as they have no CL strategy. It's pointless to use BWT and FBT to evaluate them.} \label{tb:BWT}
\setlength{\tabcolsep}{3.5mm}{
\begin{tabular}{llllll}
\toprule 
\tiny{\diagbox{Dataset}{Method$^{1}$}}   &OSFW  &ORGFW  &GEM    &CLOPS    &\textbf{\mname}  \\
\midrule 
SEPSIS    &--0.070   &--0.066 &+0.017    &+0.006    &\textbf{+0.032}     \\
COVID-19   &--0.026   &--0.015  &+0.012    &+0.004    &\textbf{+0.021}     \\
MIMIC-III   &--0.153   &--0.161 &+0.104    &+0.043    &\textbf{+0.125}     \\
USHCN     &--0.106  &--0.092    &+0.071   &+0.019  &\textbf{+0.081}   \\
\bottomrule 
\end{tabular}}
\end{table}

\begin{table}[!h]
\centering
\caption{CL Performance (FWT$\uparrow$) of Baselines.\newline \small{$^{1}$}LSTM, SR and ECEC are not listed as they have no CL strategy. It's pointless to use BWT and FBT to evaluate them. }\label{tb:FWT}
\setlength{\tabcolsep}{3.6mm}{
\begin{tabular}{llllll}
\toprule 
\tiny{\diagbox{Dataset}{Method}}   &OSFW  &ORGFW  &GEM    &CLOPS    &\textbf{\mname}  \\
\midrule 
SEPSIS    &+0.323   &+0.309  &+0.265  &+0.237    &\textbf{+0.415}     \\
COVID-19   &+0.469   &+0.478   &+0.421  &+0.289   &\textbf{+0.498}     \\
MIMIC-III   &+0.197   &+0.217  &+0.287    &+0.246   &\textbf{+0.364}     \\
USHCN     &+0.300  &+0.316    &+0.322   &+0.301  &\textbf{+0.348}   \\
\bottomrule 
\end{tabular}}
\end{table}

\begin{table*}[!h]
\centering
\footnotesize
\caption{AUC-ROC$\uparrow$, BWT$\uparrow$, FWT$\uparrow$ and Gradient Fluctuation R$\downarrow$ of Ablation of \mname}\label{tb:ablation}

\setlength{\tabcolsep}{0.6mm}{
\begin{tabular}{llccccccc}
\toprule 
 Dataset &Method$^{1}$  &Time 5   &Time 6    &Time 8   &Time 10  &BWT  &FWT &R\\
\midrule 
\multirow{2}*{SEPSIS} 
& w/o PM  &0.750±.03        &0.796±.02    &0.812±.02    &0.830±.02 &-0.102 &+0.165 &0.401\\
& w/o LM &0.743±.03        &0.790±.02    &0.801±.01    &0.825±.02 &-0.111 &+0.160 &0.400\\
&\textbf{\mname}  &\textbf{0.812±.01}     &\textbf{0.840±.01}        &\textbf{0.855±.01}    &\textbf{0.871±.01}  &\textbf{+0.030} &\textbf{+0.412} &\textbf{0.247} \\

\midrule
\multirow{2}*{COVID-19} 
& w/o PP &0.879±.02       &0.924±.02    &0.931±.01     &0.948±.01 &-0.058 &+0.195 &0.328 \\
& w/o PP &0.870±.01       &0.914±.01    &0.925±.00     &0.935±.01 &-0.088 &+0.190 &0.306 \\
&\textbf{\mname} &\textbf{0.915±.00}     &\textbf{0.919±.01}        &\textbf{0.954±.00}       &\textbf{0.964±.00}   &\textbf{+0.020}  &\textbf{+0.423} &\textbf{0.248}\\

\midrule
\multirow{3}*{MIMIC-III}
& w/o LP   &0.746±.01      &0.760±.01        &0.805±.01      &0.828±.01 &+0.053 &+0.272 &0.344 \\
& w/o PP   &0.755±.01      &0.770±.01        &0.812±.01      &0.829±.01 &+0.103 &+0.312 &0.338 \\
&\textbf{\mname} &\textbf{0.775±.01 }     &\textbf{0.784±.00}      &\textbf{0.814±.01}    &\textbf{0.840±.00} &\textbf{+0.107} &\textbf{+0.320} &\textbf{0.333}\\

\midrule
\multirow{3}*{USHCN}
& w/o LP    &0.776±.01     &0.812±.02       &0.838±.00  &0.885±.01  &+0.053 &+0.246 &0.277\\
& w/o PP    &0.775±.01     &0.810±.02       &0.840±.00          &0.886±.01  &+0.080 &+0.338 &0.249 \\
&\textbf{\mname} &\textbf{0.780±.01}     &\textbf{0.815±.00}        &\textbf{0.853±.01}     &\textbf{0.895±.00} &\textbf{+0.085} &\textbf{+0.349} &\textbf{0.246} \\
\bottomrule 
\end{tabular}
}
\end{table*}

\begin{table*}[!h]
\caption{Performance (AUC-ROC$\uparrow$, BWT$\uparrow$, FWT$\uparrow$) of \mname with Different Class Number and Training Orders}\label{tb:tasknumber}
\footnotesize
\setlength{\tabcolsep}{0.2mm}{
\begin{tabular}{l|rrrrr|rrr}
\toprule 
   &\textbf{2 Classes}    &4 Classes    &6 Classes  &8 Classes   &10 Classes   &Random &ICD-9 & Similarity\\
\midrule 
AUC-ROC    &\textbf{0.859±.01}    &0.831±.01  &0.816±.01   &0.797±.01 &0.784±.01  &0.832±.01 &0.830±.01 &\textbf{0.845±.01}\\
BWT  &\textbf{+0.153}    &+0.142  &+0.139    &+0.135 &+0.125    &+0.125 &+0.122 &\textbf{+0.133}\\
FWT &\textbf{+0.398}    &+0.384  &+0.379   &+0.365 &+0.364 &+0.364  &+0.358 &\textbf{+0.367}\\
\bottomrule 
\end{tabular}}
\end{table*}

\begin{table}[!h]
\centering
\caption{Performance (AUC-ROC$\uparrow$, BWT$\uparrow$, FWT$\uparrow$) of \mname with Different Number of Class.}\label{tb:classnumber}
\setlength{\tabcolsep}{3.3mm}{
\begin{tabular}{lrrrrr}
\toprule 
   &2    &4    &6   &8    &10    \\
\midrule 
AUC-ROC    &0.859±.01    &0.831±.01  &0.816±.01   &0.797±.01 &0.784±.01  \\
BWT  &+0.153    &+0.142  &+0.139    &+0.135 &+0.125    \\
FWT &+0.398    &+0.384  &+0.379   &+0.365 &+0.364  \\
\bottomrule 
\end{tabular}}
\end{table}

\begin{table}[!h]
\centering
\caption{Performance (AUC-ROC$\uparrow$, BWT$\uparrow$, FWT$\uparrow$) of \mname with Different Training Order in Each Time Step. }\label{tb:appendix11}
\setlength{\tabcolsep}{3.3mm}{
\begin{tabular}{lccccc}
\toprule 
 Order  &30\%  &60\%  &90\%    &BWT    &FWT  \\
\midrule 
Random    &0.757±.01    &0.788±.01  &0.832±.01   &+0.125 &+0.364  \\
ICD-9   &0.759±.01    &0.783±.01  &0.830±.01   &+0.122 &+0.358    \\
Similarity &\textbf{0.762±.01}     &\textbf{0.796±.00}  &\textbf{0.845±.01}   &\textbf{+0.133} &\textbf{+0.367} \\
\bottomrule 
\end{tabular}}
\end{table}

\begin{table*}[!h]
\caption{Classification Accuracy (AUC-ROC$\uparrow$) of \mname for Subsets with Different Data Size.\newline
\small {k\% means the volume of sub dataset is k\% of the corresponding original dataset; Bold font indicates the highest accuracy; 
* means that the accuracy of CCTS is higher 2\% than this method.}} \label{tb:appendix4}
\centering
\footnotesize
\setlength{\tabcolsep}{4mm}{
\begin{tabular}{lllllll}
\toprule 
 Dataset &Method   &20\%   &40\%    &60\%    &80\%    &100\%  \\
\midrule 
\multirow{8}*{UCR-EQ} 
& LSTM  &0.724*   &0.765*    &0.804*  &0.809*    &0.813* \\
& SR &0.758*    &0.784*  &0.828*   &0.813*   &0.831*\\
& ECEC  &0.790  &0.770*    &0.815*   &0.827* &0.838*  \\
& EWC  &0.785   &0.791*    &0.833* &0.855*     &0.862*\\
& GEM &0.780   &0.775*   &0.840*  &0.857*     &0.863* \\
& CLEAR &0.784   &0.808   &0.859    &0.864*     &0.870*\\
& CLOPS  &0.792  &0.809   &0.864 &0.871    &0.875* \\
&\textbf{CCTS}  &\textbf{0.797}     &\textbf{0.817}    &\textbf{0.872}    &\textbf{0.886}    &\textbf{0.896}   \\

\midrule 
\multirow{8}*{USHCNrain}
& LSTM   &0.701*   &0.730*   &0.732*    &0.760*  &0.763* \\
& SR    &0.731*     &0.769*  &0.782*   &0.801*  &0.805*\\
& ECEC  &0.747*     &0.774  &0.800*  &0.807*    &0.816*\\
& EWC  &0.739*     &0.768*  &0.810   &0.817    &0.826*\\
& GEM    &0.737*    &0.772   &0.809   &0.811*    &0.818*\\
& CLEAR   &0.757  &0.780 &0.812 &0.819    &0.823*\\
& CLOPS    &0.775   &0.785  &0.817   &0.825    &0.839\\
&\textbf{CCTS}    &\textbf{0.776}     &\textbf{0.790}    &\textbf{0.821}    &\textbf{0.835}   &\textbf{0.843}\\

\midrule 
\multirow{8}*{COVID-19\ } 
& LSTM  &0.713*   &0.730*    &0.765*   &0.819*    &0.834* \\
& SR   &0.751*    &0.767*   &0.806   &0.822*   &0.842* \\
& ECEC   &0.755*    &0.770* &0.796*    &0.829*  &0.856* \\
& EWC    &0.763    &0.785   &0.794*   &0.835*   &0.849*\\
& GEM    &0.769    &0.772*   &0.793*   &0.849    &0.856*\\
& CLEAR   &0.776  &0.791  &0.810    &0.856  &0.866*\\
& CLOPS &0.775    &0.789   &0.809  &0.848  &0.874\\
&\textbf{CCTS}    &\textbf{0.781}     &\textbf{0.800} &\textbf{0.821}    &\textbf{0.863}    &\textbf{0.888}   \\

\midrule 
\multirow{8}*{SEPSIS} 
& LSTM  &0.658*    &0.669*    &0.691*    &0.733*   &0.747\\
& SR &0.682    &0.700    &0.725*   &0.759*    &0.768\\
& ECEC  &0.679*    &0.702    &0.719*   &0.755*    &0.770\\
& EWC &0.685    &0.708    &0.729*   &0.768*    &0.772*\\
& GEM  &0.693    &0.704   &0.740*   &0.771*    &0.781*  \\
& CLEAR &0.687    &0.705   &0.741   &0.776    &0.789 \\
& CLOPS &0.698    &0.710   &0.745   &0.779    &0.783* \\
&\textbf{CCTS}    &\textbf{0.701}     &\textbf{0.712}    &\textbf{0.760}    &\textbf{0.794}    & \textbf{0.803}  \\
\bottomrule 
\end{tabular}
}
\end{table*}

\begin{table*}[!h]
\caption{COVID-19 Classification Accuracy with Non-uniform Training Sets and Validation Sets.\newline
\small {$\downarrow$ means the accuracy is greatly reduced.}} \label{tb:overfitting}
\centering
\footnotesize
\setlength{\tabcolsep}{4mm}{
\begin{tabular}{lllll}
\toprule 
Subset &LSTM    &SR  &ECEC    &EWC      \\
\midrule 
Male &0.955±0.013   &0.968±0.014   &0.969±0.016   &0.965±0.012 \\
Female &0.924±0.013   &0.945±0.004   &0.947±0.015   &0.939±0.018 \\
\midrule 
Age 30- &0.954±0.013   &0.965±0.014   &0.967±0.015   &0.967±0.013 \\
Age 30+ &0.923±0.014   &0.941±0.007   &0.943±0.018   &0.931±0.008$\downarrow$\\
\midrule 
Test  &0.950±0.011   &0.964±0.013   &0.968±0.015   &0.966±0.012 \\
Valid. &0.944±0.014  &0.962±0.006   &0.963±0.014   &0.954±0.003\\
\midrule 
Subset &GEM    &CLEAR &CLOPS    &\mname \\
\midrule 
Male &0.965±0.004 &0.978±0.009 &0.978±0.014   &0.971±0.010\\
Female &0.938±0.003 &0.919±0.008$\downarrow$ &0.921±0.009$\downarrow$   &0.947±0.002\\
\midrule 
Age 30- &0.964±0.009 &0.977±0.008 &0.979±0.012   &0.972±0.010\\
Age 30+  &0.923±0.040$\downarrow$ &0.902±0.006$\downarrow$ &0.914±0.007$\downarrow$   &0.945±0.006\\
\midrule 
Test &0.962±0.006 &0.979±0.009 &0.978±0.010   &0.970±0.007\\
 Valid. &0.953±0.005 &0.952±0.009$\downarrow$ &0.954±0.004$\downarrow$   &0.967±0.006\\
 \bottomrule 
\end{tabular}
}
\end{table*}

\clearpage

\begin{figure*}[!h]
\caption{Classification Accuracy Change}
\centering
\centerline{
\includegraphics[width=0.7\linewidth]{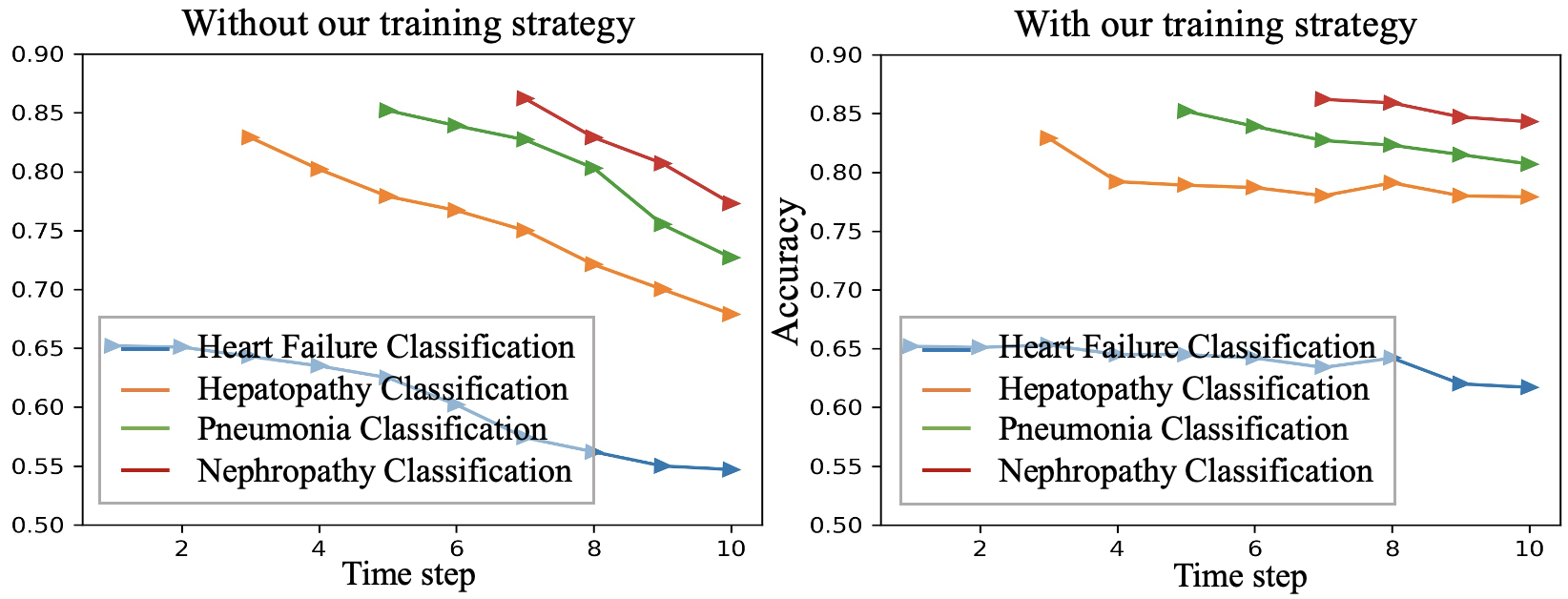}}
\label{fig:result_2}
\end{figure*}

\begin{figure*}[!h]
\centering
\caption{Multi-distribution in SEPSIS Dataset}
\includegraphics[width=0.5 \linewidth]{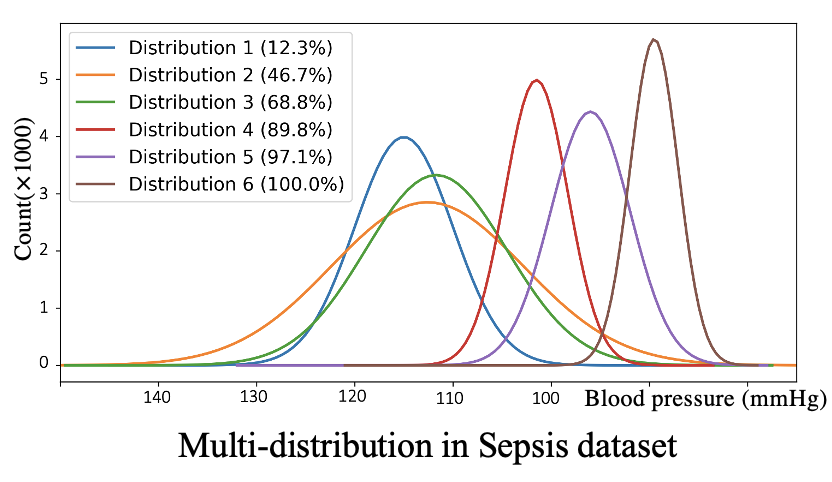}
\label{fig:distribution2}
\end{figure*}

\begin{figure*}[!h]
\centering
\caption{Multi-distribution in Datasets}
\includegraphics[width=\linewidth]{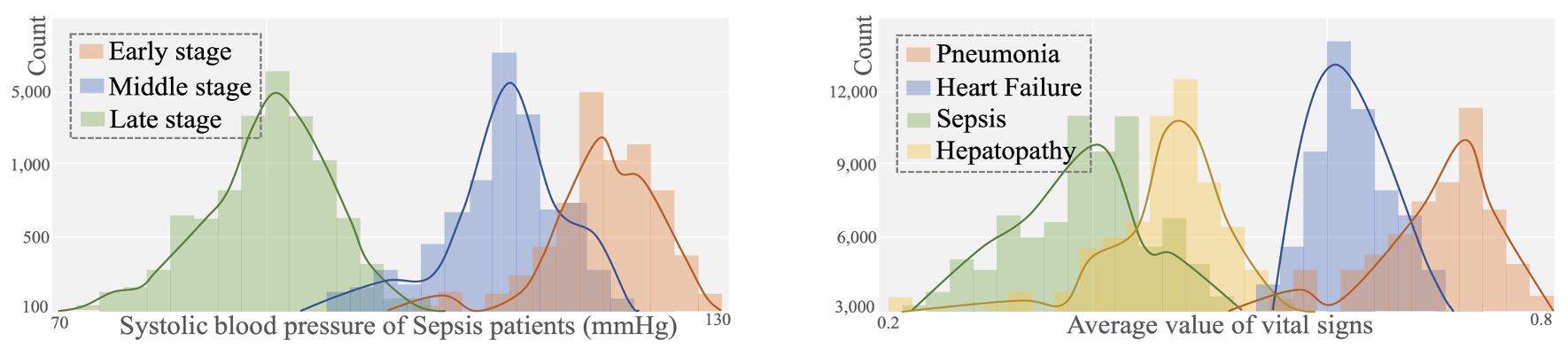}
\label{fig:distribution1}
\end{figure*}

\begin{figure*}[!h]
\centering
\caption{The Important Samples in Four SEPSIS Distribution Buffers (2,3,4,5 in Figure \ref{fig:distribution2})}
\includegraphics[width=\linewidth]{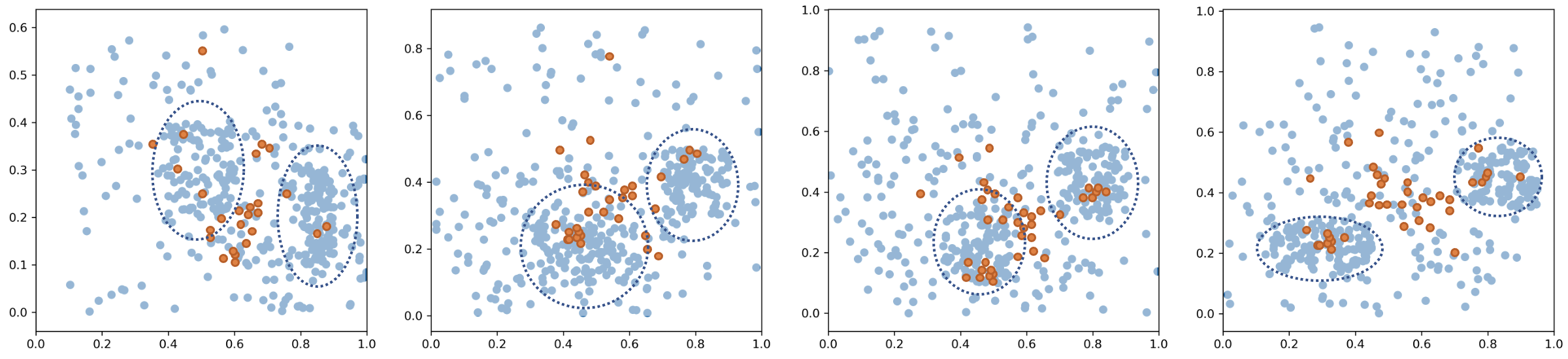}
\label{fig:distribution3}
\end{figure*}

\begin{figure*}[!h]
\caption{Parameter Test}
\centering
\includegraphics[width=0.3\linewidth]{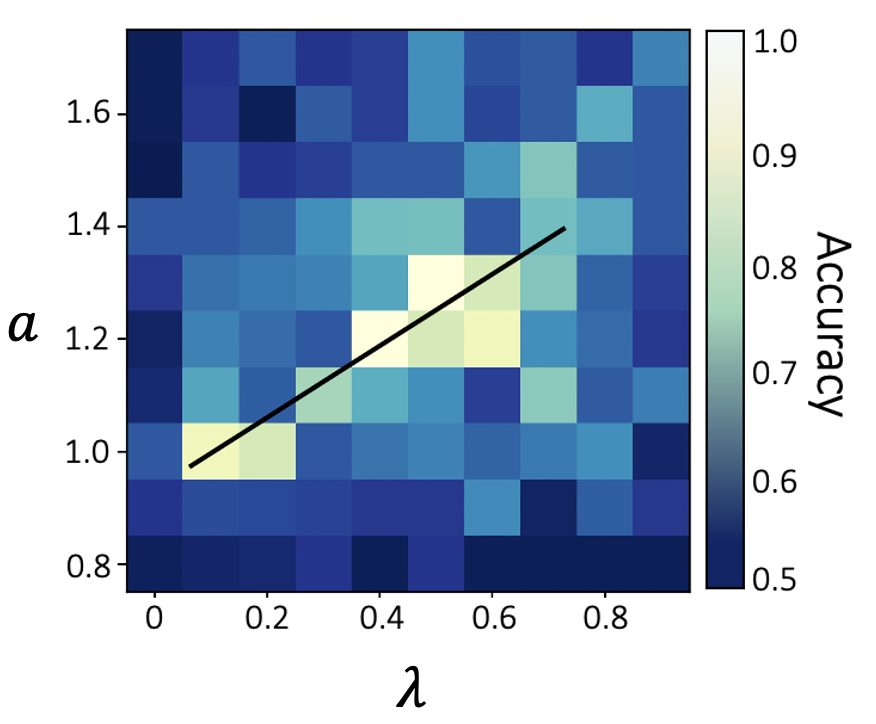}
\label{fig:appendix1}
\end{figure*}

\end{appendices}

\end{document}